\documentclass[journal]{IEEEtran}

\usepackage{cite}

\ifCLASSINFOpdf
  \usepackage[pdftex]{graphicx}
  \graphicspath{{./figures/}}

\else

\fi

\usepackage[
singlelinecheck=false % <-- important
]{caption}

\ifCLASSOPTIONcompsoc
  \usepackage[caption=false,font=normalsize,labelfont=sf,textfont=sf]{subfig}
\else
  \usepackage[caption=false,font=footnotesize]{subfig}
\fi
\usepackage{subfloat}
\usepackage{caption}
\usepackage{amsmath}
\usepackage{amssymb}
\usepackage{amstext}
\usepackage{amsfonts}
\usepackage{hyperref}
\usepackage{bbm}
\usepackage{color}

\usepackage{tablefootnote}

\usepackage{float}

\usepackage{color}
\definecolor{red}{RGB}{255,0,0}

\usepackage{multirow}

\newcommand{\layername}[1]{\texttt{#1}}

\newcommand{\revise}[1]{\textcolor{black}{#1}}
\newcommand{\secondrevise}[1]{\textcolor{black}{#1}}

\begin{document}
\title{TextBoxes++: A Single-Shot Oriented Scene Text Detector}

\author{Minghui Liao,
        \and Baoguang Shi,
        \and Xiang Bai, \textit{Senior Member, IEEE}
    \thanks{This work was supported by National Natural Science Foundation of China
(NSFC) (No. 61733007 and 61573160), the National Program for
Support of Top-notch Young Professionals and the Program
for HUST Academic Frontier Youth Team. (Corresponding author: Xiang Bai.)}
    \thanks{Minghui Liao, Baoguang Shi, Xiang Bai are with the School of Electronic Information and Communications, Huazhong University of Science and Technology (HUST), Wuhan, 430074, China. Email: \{mhliao, xbai\}@hust.edu.cn \and shibaoguang@gmail.com.}% <-this % stops a space  }
}

% \markboth{??? of \LaTeX\ Class Files,~Vol.~?, No.~?, ?~?}%
% \markboth{Revised version in color (blue)}%
% {Shell \MakeLowercase{\textit{et al.}}: Bare Demo of IEEEtran.cls for IEEE Journals}


\maketitle

\begin{abstract}
%% 1. State the problem.
%% 2. Say why it's an interesting problem.
%% 3. Say what your solution achieves.
%% 4. Say what follows from your solution.

Scene text detection is an important step of scene text recognition system and also a challenging problem. Different from general object detection, the main challenges of scene text detection lie on arbitrary orientations, small sizes, and significantly variant aspect ratios of text in natural images. In this paper, we present an end-to-end trainable fast scene text detector, named TextBoxes++, which detects arbitrary-oriented scene text with both high accuracy and efficiency in a single network forward pass. No post-processing other than an efficient non-maximum suppression is involved. We have evaluated the proposed TextBoxes++ on four public datasets. In all experiments, TextBoxes++ outperforms competing methods in terms of text localization accuracy and runtime. More specifically, TextBoxes++ achieves an f-measure of 0.817 at 11.6fps for $1024 \times 1024$ ICDAR 2015 Incidental text images, and an f-measure of 0.5591 at 19.8fps for $768 \times 768$ COCO-Text images. Furthermore, combined with a text recognizer, TextBoxes++ significantly outperforms the state-of-the-art approaches for word spotting and end-to-end text recognition tasks on popular benchmarks. Code is available at: \href{https://github.com/MhLiao/TextBoxes\_plusplus}{https://github.com/MhLiao/TextBoxes\_plusplus}.

\end{abstract}

\IEEEpeerreviewmaketitle

\section{Introduction}
Scene text is one of the most general visual objects in natural scenes, which frequently appears on road signs, license plates, product packages, \emph{etc}. Reading scene text facilitates a lot of useful applications, such as image-based geolocation. Some applications of scene text detection and recognition are~\cite{yi2014scene,xiong2016text,aaai/KangKY17,RongYT16}. Despite the similarity to traditional OCR, scene text reading is much more challenging due to large variations in both foreground text and background objects, arbitrary orientations, aspect ratios, as well as uncontrollable lighting conditions, \emph{etc}., as summarized in~\cite{ye2015text}. Owing to these inevitable challenges and complexities, traditional text detection methods tend to involve multiple processing steps, \emph{e.g.} character/word candidate generation~\cite{Pan2011,neumann2012real,jaderberg2016reading}, candidate filtering~\cite{jaderberg2016reading}, and grouping~\cite{bai2013scene}. They often end up struggling to get each module work properly, requiring much effort in tuning parameters and designing heuristic rules, also slowing down detection speed. 

\begin{figure*}
\centering
\includegraphics[width=1.0\linewidth]{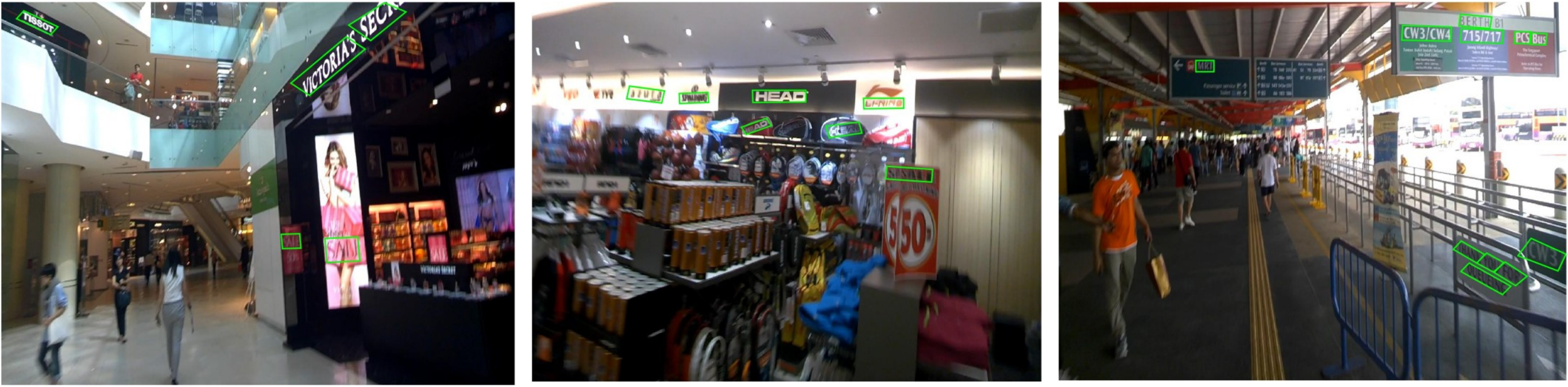}
\caption{Detection results on some challenging images.}
\label{fig:exampleresults}
% \vspace*{-5mm}
\end{figure*}

Inspired by the recent developments in object detection~\cite{liu2015ssd,ren2015faster}, we propose to detect text by \emph{directly} predicting word bounding boxes with quadrilaterals via a single neural network that is end-to-end trainable. We call it TextBoxes++. Concretely, we replace the rectangular box representation in conventional object detector by a quadrilateral or oriented rectangle representation. Furthermore, to achieve better receptive field that covers text regions which are usually long, we design some ``long'' convolutional kernels to predict the bounding boxes. TextBoxes++ directly outputs word bounding boxes at multiple layers by jointly predicting text presence and coordinate offsets to \emph{anchor boxes}~\cite{ren2015faster}, also known as \emph{default boxes}~\cite{liu2015ssd}. The final outputs are the non-maximum suppression outputs on all boxes. A single forward pass in the network densely detects multi-scale text boxes all over the image. As a result, the detector has a great advantage in speed. We will show by experiments that TextBoxes++ achieves both high accuracy and high speed with a single forward pass on single-scale inputs, and even higher accuracy when performing multiple passes on multi-scale inputs. Some text detection examples on several challenging images are depicted in Fig.~\ref{fig:exampleresults}.

We further combine TextBoxes++ with CRNN~\cite{shi2015end}, an open-source text recognition module. The recognizer not only produces extra recognition outputs but also regularizes text detection with its semantic-level awareness, thus further boosts the accuracy of word spotting considerably. The combination of TextBoxes++ and CRNN yields state-of-the-art performance on both word spotting and end-to-end text recognition tasks, which appears to be a simple yet effective solution to robust text reading in the wild. 

% The performance of TextBoxes++ is further boosted with a text recognition module.
% Furthermore, we argue that word recognition is helpful to distinguish text
% from backgrounds, especially when words are confined to a given set, \emph{i.e.} a lexicon. We adopt a successful text recognition algorithm, CRNN~\cite{shi2015end}, in conjunction with TextBoxes++. The text recognizer not only provides extra recognition outputs but also regularizes text detection with its semantic-level awareness, thus further boosting the accuracy of word spotting considerably. The combination of TextBoxes++ and CRNN yields the state-of-the-art performance on word spotting and end-to-end text recognition tasks, which appears to be a simple yet effective solution to robust text reading in the wild. 

% As far as we know, most of the scene text detectors are evaluated by an IOU threshold, which is inherited from object detection.  However, text detection is slightly different from object detection. It is meaningless to only detect text without recognizing them. In other words, the final purpose of text detection is to recognize them. An outstanding text detector should not only perform well on IOU evaluation but also get leading performance on the task of word spotting and end-to-end recognition. This is why we combine TextBoxes++ with a text recognizer to perform word spotting and end-to-end recognition. The state-of-the-art performances of word spotting and end-to-end recognition further verify the robustness of TextBoxes++.

A preliminary version of this study called TextBoxes was exposed in~\cite{LiaoSBWL17}. \revise{\secondrevise{The current paper is an extension of~\cite{LiaoSBWL17}, which extends TextBoxes with four main improvements}: 1) We extend TextBoxes, a horizontal text detector, to a detector that can handle arbitrary-oriented text; 2) We revisit and improve the network structure and the training process, which leads to a further performance boost; 3) More comparative experiments have been conducted to further demonstrate the efficiency of TextBoxes++ in detecting arbitrary-oriented text in natural images; 4) We refine the combination of detection and recognition by proposing a novel score which elegantly utilizes both the information of detection and recognition.}

\revise{The main contributions of this paper are three folded: 1) The proposed arbitrary-oriented text detector TextBoxes++ is accurate, fast, and end-to-end trainable. As compared to closely related methods, TextBoxes++ has a much simpler yet effective pipeline. 2) This paper also offers some comparative studies on some important design choices and other issues, including bounding box representations, model configurations, and the effect of different evaluation methods. The conclusions of these studies may generalize to other text reading algorithms and give insights on some important issues. 
% 3) We have conducted extensive experiments which show that TextBoxes++ achieves both high accuracy and high speed with a single forward pass on single-scale inputs, and even higher accuracy when performing multiple passes on multi-scale inputs. \textcolor{red}{All these results show that TextBoxes++ significantly advance the state-of-the-arts in text detection.}
3) We also introduce the idea of using recognition results to further refine the detection results thanks to the semantic-level awareness of recognized text. To the best of our knowledge, this intuitive yet effective combination has not been exploited before.
%We argue that recognition is able to refine the detection results and discuss the effect of recognition on two benchmarks.
}

% The contributions of this paper are three-fold:
% \begin{enumerate}
% \item An end-to-end trainable oriented text detector is proposed. The detector is both accurate and fast.
% \item 
% \item Second,  we argue that IOU which is inherited from object detection is not accurate for evaluate the performance of a text detector and propose a word spotting/end-to-end recognition framework that effectively combines detection and recognition. 
% \item Finally,  we compare and analyze the bounding boxes representation of rotated rectangle and quadrilateral. 
% \end{enumerate}

% \begin{enumerate}
% \item We extended TextBoxes, 
% \item We 
% \item We made experimental discussions on several important design choices, including bounding box representation and evaluation methods. The conclusions may generalize to other text reading algorithms.
% \end{enumerate}

The rest of the paper is organized as follows. Section~\ref{sec: related works} briefly reviews some related works on object detection and scene text detection. The proposed method is detailed in Section~\ref{sec: method description}. We present in Section~\ref{sec: experimental results} some experimental results. \revise{A detailed comparison with some closely related methods is given in Section~\ref{sec:detailedcomparison}}. Finally, conclusions are drawn in Section~\ref{sec: conclusion}.

\section{Related Works} \label{sec: related works}
\subsection{Object detection}
Object detection is a fundamental problem in computer vision, which aims to detect general objects in images. Recently, there are two mainstream CNN-based methods on this topic: R-CNN based methods~\cite{rcnn,fast_rcnn,ren2015faster} and YOLO-based methods~\cite{RedmonDGF15,liu2015ssd}. %The later approaches directly regress bounding boxes in parallel on the whole images. 

\paragraph{R-CNN based object detection} R-CNN~\cite{rcnn} views a detection problem as a classification problem leveraging the development of classification using convolutional neural networks(CNN). It first generates proposals by selective search~\cite{selective-search} and then feeds the proposals into a CNN to extract deep features based on which an SVM~\cite{svm} is applied for classification. Fast R-CNN~\cite{fast_rcnn} improves R-CNN by extracting deep features of the proposals from the feature maps via RoI pooling~\cite{fast_rcnn} instead of cropping from the origin image. This significantly simplifies the pipeline of R-CNN. Furthermore, a regression branch is also used in fast R-CNN to get more accurate bounding boxes. Faster R-CNN~\cite{ren2015faster} further improves the speed of fast R-CNN~\cite{fast_rcnn} by introducing an end-to-end trainable region proposal network based on anchor boxes to generate proposals instead of using selective search. The generated proposals are then fed into a Fast R-CNN network for detection.
% Recently, two other R-CNN based methods have been proposed: R-FCN~\cite{rfcn} focusing on improving speed and Mask R-CNN~\cite{mask-rcnn} paying more attention to performance.

\paragraph{YOLO-based object detection} The original YOLO~\cite{RedmonDGF15} directly regresses the bounding boxes on the feature maps of the whole image. A convolutional layer is used to predict the bounding boxes on different areas instead of the RPN and RoI pooling used in~\cite{ren2015faster}. This results in a significantly reduced runtime. Another YOLO-based method is SSD~\cite{liu2015ssd} which predicts object bounding boxes using default boxes of different aspect ratios on different stages. The concept of default boxes~\cite{liu2015ssd} is similar to anchor boxes~\cite{ren2015faster}, which are fixed as the reference systems of the corresponding ground truths. The translation and scale invariances for regression are achieved by using default boxes of different aspect ratios and scales at each location, which eases the regression problem. SSD further improves the performance of original YOLO while maintaining its runtime. 
% There are also some other variants of YOLO-based methods. Two representative examples are YOLO9000~\cite{yolo9000} and DSSD~\cite{dssd}. Both of them aim at improving the performance of previous works at an acceptable loss of speed.

\subsection{Text detection}
\label{subsubsec:textdetection}
A scene text reading system is usually composed of two main components: text detection and text recognition. The former component localizes text in images mostly in the form of word bounding boxes. The latter one transcripts cropped word images into machine-interpretable character sequences. In this paper, we cover both aspects, but more attention is paid to detection. In general, most text detectors can be roughly classified into several categories following two classification strategies based on primitive detection targets and shape of target bounding boxes, respectively.

\paragraph{Classification strategy based on primitive detection targets} Most text detection methods can be roughly categorized into three categories:\\
\emph{1) Character-based}: Individual characters or parts of the text are first detected and then grouped into words~\cite{neumann2012real}. A representative example is the method proposed in~\cite{neumann2012real} which locates characters by classifying Extremal Regions and then groups the detected characters by an exhaustive search method. Other popular examples of this type are the works in~\cite{Yao2012,li2014characterness,huang2014robust,gomez2013multi,text-line-detection-based};\\
\emph{2) Word-based}: Words are directly extracted in the similar manner as general object detection \cite{ZhaoLK10,jaderberg2016reading,Zhong2016,TextProposal}. In the representative work~\cite{jaderberg2016reading}, the authors propose an R-CNN-based~\cite{rcnn} framework, where word candidates are first generated with class-agnostic proposal generators followed by a random forest classifier. Then a convolutional neural network for bounding box regression is adopted to refine the bounding boxes. YOLO network~\cite{RedmonDGF15} is used in \cite{gupta2016synthetic} which also relies on a classifier and a regression step to remove some false positives;\\
\emph{3) Text-line-based}: Text lines are detected and then broken into words. The works in~\cite{huang2014robust,zhang2015symmetry,long2015fully,Zhang_2016_CVPR} are such examples. In~\cite{zhang2015symmetry}, the authors propose to detect text lines making use of the symmetric characteristics of text. This idea is further exploited in~\cite{long2015fully} by using a fully convolutional network to localize text lines.

\paragraph{Classification strategy based on the shape of target bounding boxes} Following this classification strategy, the text detection methods can be categorized into two categories:\\
\emph{1) Horizontal or nearly horizontal}: These methods focus on detecting horizontal or nearly horizontal text in images. An algorithm based on AdaBoost is proposed in~\cite{cvpr/ChenY04}. Then, Yi et al.~\cite{tip/YiT12} propose a 3-stage framework which consists of boundary clustering, stroke segmentation, and string fragment classification. Some examples which are inspired by object detection methods are~\cite{jaderberg2016reading,gupta2016synthetic}. They all use horizontal rectangle bounding boxes as predict targets, which is very similar to general object detection methods. Another popular method of this type is the work in~\cite{eccv/TianHHH016} which detects nearly horizontal text parts and then links them together to form word candidates. Besides, Cao et al.~\cite{tip/CaoRZGF15} try to use deblurring techniques for more robust detection results.\\
\emph{2) Multi-oriented}: As compared to horizontal or nearly horizontal detection, multi-oriented text detection is more robust because scene text can be in arbitrary orientations in images. There exist several works which attempt to detect multi-oriented text in images. In~\cite{Yao2012}, the authors propose two sets of rotation-invariant features for detecting multi-oriented text. The first set is component level features such as estimated center, scale, direction before feature computation. The second one is chain level features such as size variation, color self-similarity, and structure self-similarity. Kang et al.~\cite{KangLD14} build a graph of MSERs~\cite{matas2004robust} followed by a higher-order correlation clustering to generate multi-oriented candidates. A unified framework for multi-oriented text detection and recognition is proposed in~\cite{tip/YaoBL14}. They use the same features for text detection and recognition. Finally, a texton-based texture classifier is used to discriminate text and no-text candidates. In~\cite{Zhang_2016_CVPR,corr/YaoBSZZC16}, multi-oriented text bounding boxes are generated from text saliency maps given by a dedicated segmentation network. Recently, Shi et al.~\cite{corr/ShiBB17} propose to detect text with segments and links. More precisely, they first detect a number of text parts named segments and meanwhile predict the linking relationships among neighboring segments. Then related segments are linked together to form text bounding boxes. A U-shape fully convolutional network is used in~\cite{corr/EAST} for detecting multi-oriented text. In this work, the authors also introduce a PVANet~\cite{pvanet} for efficiency. Quadrilateral sliding windows, a Monte-Carlo method, and a smooth Ln loss are proposed in~\cite{LiuJ17b} to detect oriented text, which is effective while complicated.

\subsection{TextBoxes++ Versus some related works}
TextBoxes++ is a \emph{word-based} and \emph{multi-oriented} text detector. In contrast to~\cite{jaderberg2016reading}, which consists of three detection steps and each step includes more than one algorithm, TextBoxes++ has a much simpler pipeline. Only an end-to-end training of one network is required. As described in Section~\ref{subsubsec:textdetection}, Tian et al.~\cite{eccv/TianHHH016} and Shi et al.~\cite{corr/ShiBB17} propose to detect text parts and then link them together. Both of them achieve impressive results for long words. However, the proposed method in~\cite{eccv/TianHHH016} has limited adaptability for oriented text due to the single orientation of the Bidirectional Long Short-Term Memory (BLSTM)~\cite{blstm}, and the work in~\cite{corr/ShiBB17} has two super parameters for linking the segments, which are determined by grid search and difficult to adjust. 
% Zhou et al.~\cite{corr/EAST} adopt a U-shape network with deconvolutional layers to regress rotated boxes or quadrilaterals. Nevertheless, the deconvolutional layers are time consuming. So a simple backbone is required to speed up their network, which leads to poorer performance. When the same backbone (\textit{e.g.} VGG network) is used for both~\cite{corr/EAST} and TextBoxes++, TextBoxes++ is much faster thanks to its fully convolutional architecture .
% \revise{Zhou et al.~\cite{corr/EAST} first generates a score map (text region segmentation) by a U-shape network~\cite{u-net} and then regresses the oriented rectangles or quadrilaterals based on the generated score map. In this way, it needs pyramid-like deconvolutional layers for accurate segmentation. However, TextBoxes++ directly classifies and regresses the default boxes on the convolutional feature maps, which is simpler yet effective. When the same backbone (\textit{e.g.} VGG network) is used for both~\cite{corr/EAST} and TextBoxes++, TextBoxes++ is faster thanks to its fully convolutional architecture.}
\revise{The method in~\cite{corr/EAST} is considered as the current state-of-the-art approach. It relies on a U-shape network to simultaneously generate a score map for text segmentation and the bounding boxes. Yet, an accurate text region segmentation is challenging in itself. Besides, The extra pyramid-like deconvolutional layers involved in text region segmentation require additional computation. Whereas, TextBoxes++ directly classifies and regresses the default boxes on the convolutional feature maps, where richer information is reserved as compared to the segmentation score map. TextBoxes++ is thus much simpler, avoiding the time consuming on pyramid-like deconvolution.}

\revise{One of} the most related work to TextBoxes++ is SSD~\cite{liu2015ssd}, a recent development in object detection. Indeed, TextBoxes++ is inspired by SSD~\cite{liu2015ssd}. The original SSD aims to detect general objects in images but fails on words having extreme aspect ratios. TextBoxes++ relies on specifically designed text-box layers to efficiently solve this problem. This results in a significant performance improvement. Furthermore, SSD can only generate bounding boxes in terms of horizontal rectangles, while TextBoxes++ can generate arbitrarily oriented bounding boxes in terms of oriented rectangles or general quadrilaterals to deal with oriented text.

\revise{
Another closely related work to TextBoxes++ is the method in~\cite{LiuJ17b}, which proposes quadrilateral sliding windows and a Monte-Carlo method for detecting oriented text. In TextBoxes++, we use horizontal rectangles as default boxes, and hence have much fewer default boxes in every region. Benefiting from the horizontal rectangle default boxes, TextBoxes++ also enjoys a much simpler strategy for matching default boxes. Moreover, TextBoxes++ simultaneously regresses the maximum horizontal rectangles of the bounding boxes and the quadrilateral bounding boxes, which makes the training more stable. Furthermore, we propose a novel score by combining the detection and recognition scores to further refine the detection results.}

% \revise{Another related work to TextBoxes++ is \cite{LiuJ17b}. The main differences between TextBoxes++ and \cite{LiuJ17b} are in three dimensions: 1) TextBoxes++ uses horizontal rectangles as default boxes instead of rectangles with different angles. We argue that the receptive fields of the convolutional feature map are all in horizontal rectangles, so it is no need to apply oriented rectangle default boxes in general scene text detection. 2) Benefiting from the horizontal rectangle default boxes, TextBoxes++ enjoys a much simpler matching strategies for default boxes and ground truth, using the maximum horizontal rectangles instead of quadrilaterals. As we all know, computing the coverage between two horizontal rectangles is much easier than computing it between two general quadrilaterals, even a Monte-Carlo method is used. 3) TextBoxes++ uses a simpler smooth-L1 loss function instead of a smooth L$n$ loss proposed in~\cite{LiuJ17b}.}

The ultimate goal of text detection is to spot words or recognize text in images. On the other hand, the semantic-level awareness of spotted words or recognized words can also help to further regularize text detection results. Following this idea, we propose to combine a text recognizer with TextBoxes++ for word spotting and end-to-end recognition, and use the confidence scores of recognized words to regularize the detection outputs of TextBoxes++. For that, we adopt a state-of-the-art text recognizer called CRNN~\cite{shi2015end}, which directly outputs character sequences given input images and is also end-to-end trainable. Other text recognizers such as~\cite{jaderberg2016reading} are also applicable. Such simple pipeline for word spotting and end-to-end recognition is very different from the classical pipelines.

%Note that it is also possible to adopt other recognizers, such as~\cite{jaderberg2016reading}.

\section{Detecting oriented text with TextBoxes++}\label{sec: method description}

\begin{figure*}[!ht]
\begin{center}
\includegraphics[width=1.0\linewidth]{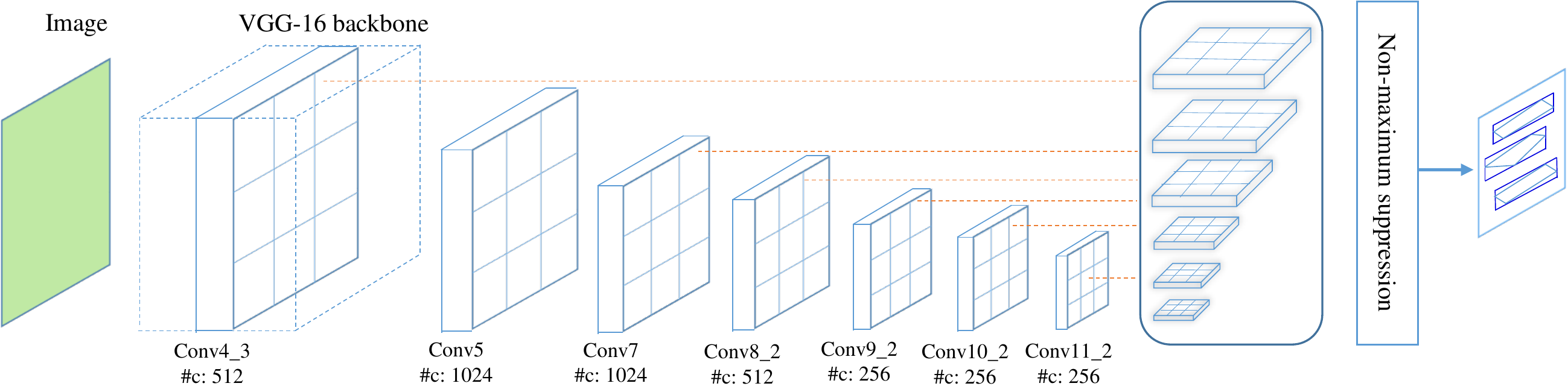}
\end{center}
\caption{The architecture of TextBoxes++, a fully convolutional network including 13 layers from VGG-16 followed by 10 extra convolutional layers, and 6 Text-box layers connected to 6 intermediate convolutional layers. Each location of a text-box layer predicts an n-dimensional vector for each default box consisting of the text presence scores (2 dimensions), horizontal bounding rectangles offsets (4 dimensions), and rotated rectangle bounding box offsets (5 dimensions) or quadrilateral bounding box offsets (8 dimensions). A non-maximum suppression is applied during test phase to merge the results of all 6 text-box layers. \revise{Note that ``\#c" stands for the number of channels.}}
\label{fig:pipeline}
% \vspace{-5mm}
\end{figure*}

\subsection{Overview}
TextBoxes++ relies on an end-to-end trainable fully convolutional neural network to detect arbitrary-oriented text. The basic idea is inspired by an object detection algorithm SSD proposed in~\cite{liu2015ssd}. We propose some special designs for adapting SSD network to efficiently detect oriented text in natural images. More specifically, we propose to represent arbitrary-oriented text by quadrilaterals~\cite{corr/EAST,HeZWT17} or oriented rectangles~\cite{r2cnn}. Then we adapt the network to predict the regressions from default boxes to oriented text represented by quadrilaterals or oriented rectangles. To better cover the text which could be dense in some area, we propose to densify default boxes with vertical offsets. Furthermore, we adapt the convolution kernels to better handle text lines which are usually long objects as compared to general object detection. These network adaptions are detailed in Section~\ref{subsec:architecture}. Some specific training adaptions for arbitrary-oriented text detection are presented in Section~\ref{subsec:training}, including on-line hard negative mining and data augmentation by a novel random cropping strategy specifically designed for text which is usually small. TextBoxes++ detects oriented text at 6 different scales in 6 stages. During the test phase, the multi-stage detection results are merged together by an efficient cascaded non-maximum suppression based on IOU threshold of quadrilaterals or oriented rectangles (see Section~\ref{subsec:testing}). Finally, we also propose an intuitive yet effective idea of using text recognition to further refine detection results thanks to the semantic-level awareness of recognized text. This is discussed in Section~\ref{subsec:word-spotting}.

\subsection{Proposed network}
\label{subsec:architecture}

\subsubsection{Network architecture}
The architecture of TextBoxes++ is depicted in Fig.~\ref{fig:pipeline}. It inherits the popular VGG-16 architecture~\cite{simonyan2014very}, keeping the layers from \layername{conv1\_1} through \layername{conv5\_3}, and converting the last two fully-connected layers of VGG-16 into convolutional layers (\layername{conv6} and \layername{conv7}) by parameters down-sampling~\cite{liu2015ssd}. Another eight convolutional layers divided into four stages (\layername{conv8} to \layername{conv11}) with different resolutions by max-pooling are appended after \layername{conv7}. Multiple output layers, which we call \emph{text-box layers}, are inserted after the last and some intermediate convolutional layers. They are also convolutional layers to predict outputs for aggregation and then undergo an efficient non-maximum suppression (NMS) process. Putting all above together, TextBoxes++ is a fully-convolutional structure consisting of only convolutional and pooling layers. As a result, TextBoxes++ can adapt to images of arbitrary size in both training and testing phases. Compared with a preliminary study in~\cite{LiaoSBWL17} of this paper, TextBoxes++ replaces the last global average pooling layer with a convolutional layer, which is furthermore beneficial for multi-scale training and testing.

\subsubsection{Default boxes with vertical offsets}
\begin{figure}[!htbp]
\begin{center}
\includegraphics[width=0.9\linewidth]{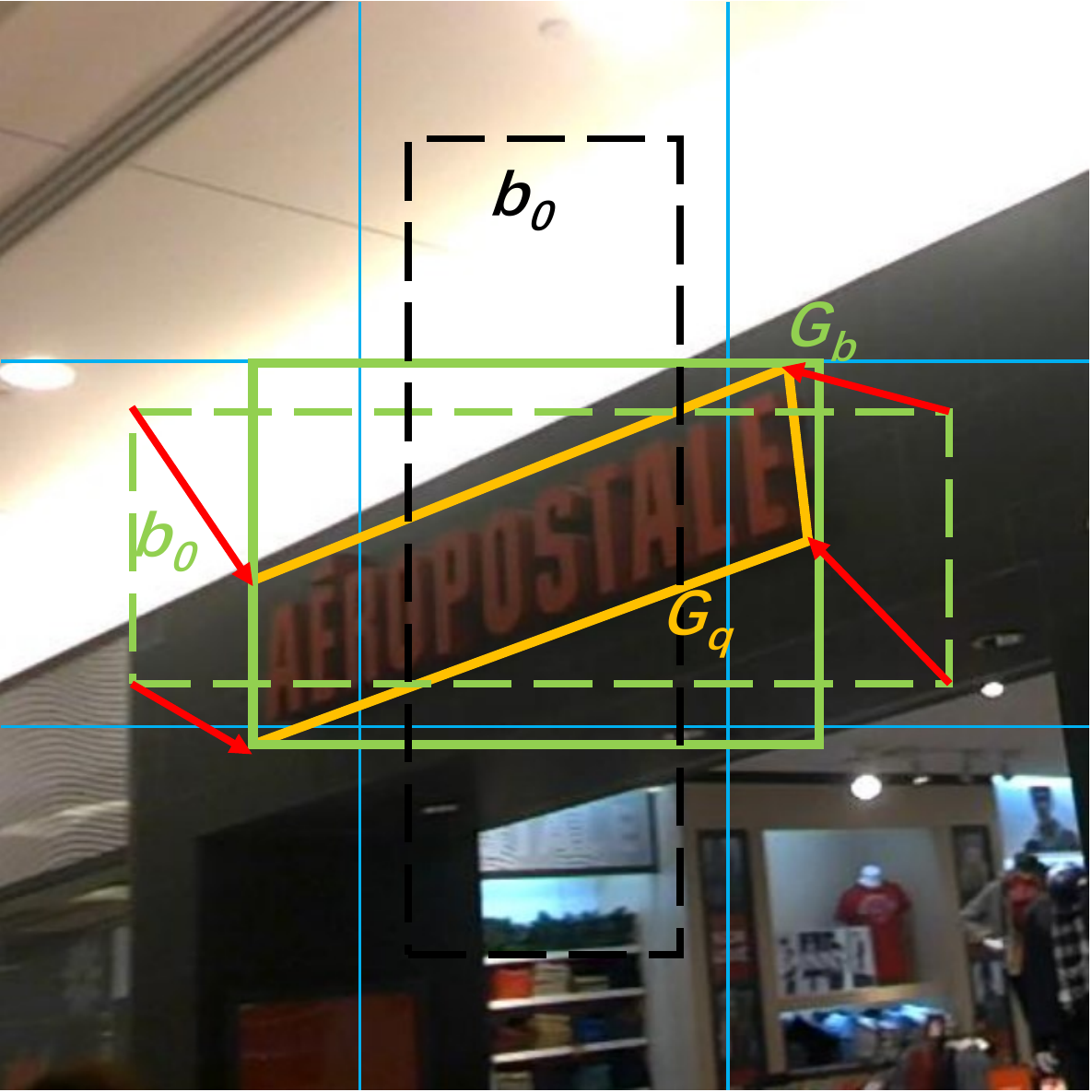}
\end{center}
% \vspace{-5mm}
\caption{Illustration of the regression (red arrows) from a matched default box (green dashed) to a ground truth target quadrilateral (yellow) on a $3 \times 3$ grid. Note that the black dashed default box is not matched to the ground truth. The regression from the matched default box to the minimum horizontal rectangle (green solid) containing the ground truth quadrilateral is not shown for a better visualization.}
\label{fig:default-boxes}
% \vspace{-2mm}
\end{figure}

Text-box layers are the key component of TextBoxes++. A text-box layer simultaneously predicts text presence and bounding boxes, conditioned on its input feature maps. The output bounding boxes of TextBoxes++ include oriented bounding boxes $\{\mathbf{q}\}$ or $\{\mathbf{r}\}$, and minimum horizontal bounding rectangles $\{\mathbf{b}\}$ containing the corresponding oriented bounding boxes. This is achieved by predicting the regression of offsets from a number of pre-designed horizontal default boxes at each location (see Fig.~\ref{fig:default-boxes} for an example). More precisely, let $\mathbf{b}_0=(x_0, y_0, w_0, h_0)$ denote a horizontal default box, which can also be written as $\mathbf{q}_0=(x^q_{01}, y^q_{01}, x^q_{02}, y^q_{02}, x^q_{03}, y^q_{03}, x^q_{04}, y^q_{04})$ or $\mathbf{r}_0=(x^r_{01}, y^r_{01}, x^r_{02}, y^r_{02}, h^r_0)$, where $(x_0 , y_0)$ means the center point of a default box and $w_0$ and $h_0$ are the width and height of a default box respectively. The relationships among $\mathbf{q}_0$, $\mathbf{r}_0$ and $\mathbf{b}_0$ are as following:
\begin{align}
\begin{split}
 x^q_{01} &=x_0 - w_0 / 2,  y^q_{01} =y_0 - h_0 / 2,\\
 x^q_{02} &=x_0 + w_0 / 2,  y^q_{02} =y_0 - h_0 / 2,\\
 x^q_{03} &=x_0 + w_0 / 2,  y^q_{03} =y_0 + h_0 / 2,\\
 x^q_{04} &=x_0 - w_0 / 2,  y^q_{04} =y_0 + h_0 / 2,\\
 x^r_{01} &=x_0 - w_0 / 2,  y^r_{01} =y_0 - h_0 / 2,\\
 x^r_{02} &=x_0 + w_0 / 2,  y^r_{02} =y_0 - h_0 / 2,\\
 h^r_0 &=h_0.
\end{split}
\label{eq:decode-default box}
\end{align}
At each map location, it outputs the classification score and offsets to each associated default box denoted as $q_0$ or $r_0$ in a convolutional manner. For the quadrilateral representation of oriented text, the text-box layers predict the values of $(\Delta x, \Delta y, \Delta w, \Delta h,\Delta x_1, \Delta y_1, \Delta x_2, \Delta y_2,\Delta x_3, \Delta y_3, \Delta x_4, c)$, indicating that a horizontal rectangle $\mathbf{b}=(x, y, w, h)$ and a quadrilateral $\mathbf{q}=(x^q_{1}, y^q_{1}, x^q_{2}, y^q_{2},x^q_{3}, y^q_{3}, x^q_{4}, y^q_{4})$ given in the following are detected with confidence $c$:
\begin{align}
\begin{split}
  x &= x_0 + w_0 \Delta x, \\
  y &= y_0 + h_0 \Delta y, \\
  w &= w_0 \exp(\Delta w), \\
  h &= h_0 \exp(\Delta h),\\
  x^q_{n} &= x^q_{0n} + w_0\Delta x^q_{n}, {n = 1, 2, 3, 4}  \\
  y^q_{n}&= y^q_{0n} + h_0\Delta y^q_{n}, {n = 1, 2, 3, 4}.
\end{split}
\label{eq:decode-polygon}
\end{align}
When the representation by rotated rectangles is used, the text-box layers predict the value of  $(\Delta x, \Delta y, \Delta w, \Delta h, \Delta x_1, \Delta y_1, \Delta x_2, \Delta y_2,\Delta h^r, c)$, and the rotated rectangle $\mathbf{r}=(x^r_{1}, y^r_{1}, x^r_{2}, y^r_{2}, h^r)$ is calculated as following:
\begin{align}
\begin{split}
  x^r_{n} &= x^r_{0n} + w_0\Delta x^r_{n}, {n = 1, 2} \\
  y^r_{n} &= y^r_{0n} + h_0\Delta y^r_{n}, {n = 1, 2}\\
  h^r &= h^r_0 \exp(\Delta h_r).
\end{split}
\label{eq:decode-rox}
\end{align}

In the training phase, ground-truth word boxes are matched to default boxes according to box overlap following the matching scheme in~\cite{liu2015ssd}. As shown in Fig.~\ref{fig:default-boxes}, the minimum bounding horizontal rectangle of a rotated rectangle or a quadrilateral is used to match the default boxes for efficiency. Note that there are a number of default boxes with different aspect ratios at each location. In this way, we effectively divide words by their aspect ratios, allowing TextBoxes++ to learn specific regression and classification weights that handle words of similar aspect ratio. Therefore, the design of default boxes is highly task-specific.

\begin{figure}[!htbp]
\begin{center}
\captionsetup[subfigure]{justification=centering}
    \centering
\subfloat[\label{subfig-1:denser-box}]{%
       \includegraphics[width=0.23\textwidth]{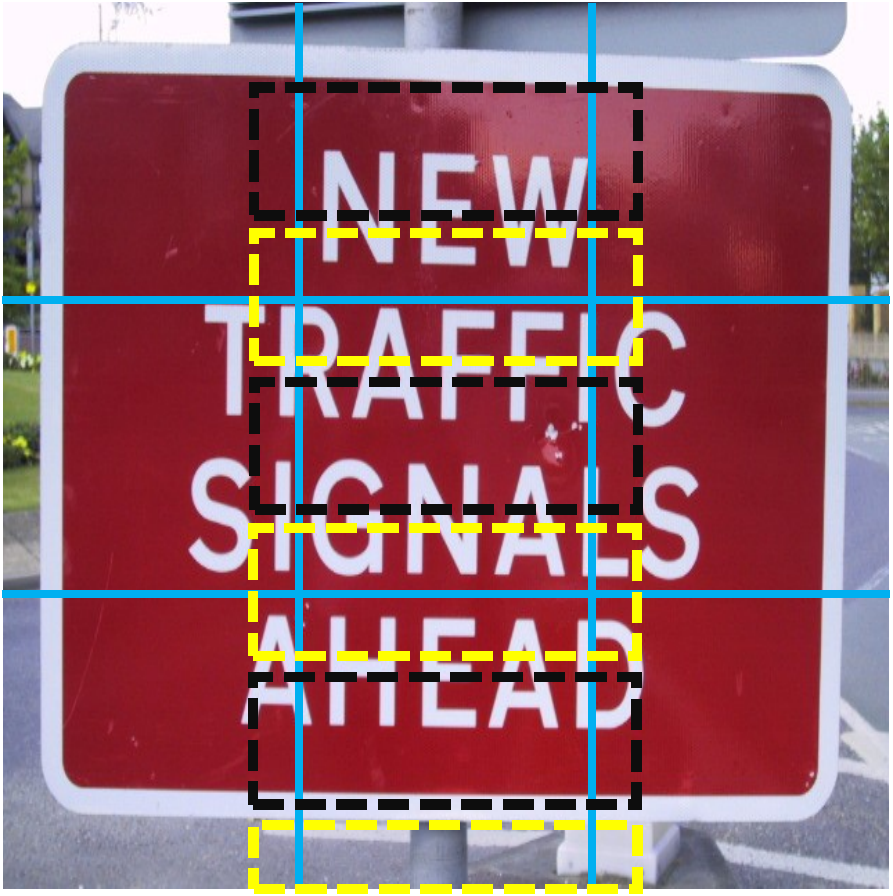}
     }
\subfloat[\label{subfig-2:denser-box}]{%
       \includegraphics[width=0.23\textwidth]{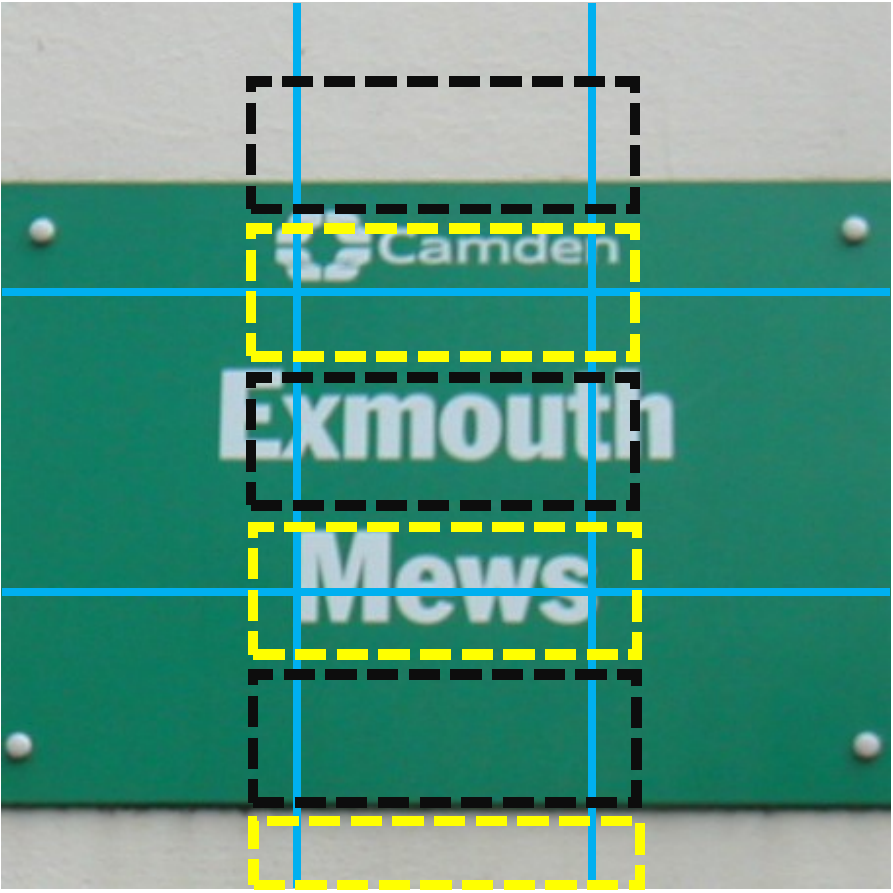}
     }
\end{center}
 \caption{Vertical offsets of default boxes on a $3 \times3 $ grid. Black (\textit{resp.} yellow) dashed bounding boxes are normal default boxes (\textit{resp.} default boxes with vertical offsets). Note that only the default boxes of appropriate aspect ratio are shown for better visualization.}
\label{fig:denser-box}
\vspace{-3mm}
\end{figure}

Different from general objects, words tend to have large aspect ratios. Therefore, the preliminary study TextBoxes in~\cite{LiaoSBWL17} include ''long'' default boxes that have large aspect ratios. Specifically, for the horizontal text dataset, we defined 6 aspect ratios for default boxes, including $1$,$2$,$3$,$5$,$7$, and $10$. However, TextBoxes++ aims to detect arbitrary-oriented text. Consequently, we set the aspect ratios of default boxes to $1, 2, 3, 5, 1/2, 1/3, 1/5$. Furthermore, text is usually dense on a certain area, so each default box is set with a vertical offset to better cover all text, which makes the default boxes dense in the vertical orientation. An example is shown in Fig.~\ref{fig:denser-box}. In Fig.~\ref{fig:denser-box}(a), the normal default box (black dashed) in the middle can not handle the two words close to it at the same time. In this case, one word would be missed for detection if no vertical offset is applied. In Fig.~\ref{fig:denser-box}(b), the normal default boxes does not cover the bottom word at all, which also demonstrates the necessity of using default boxes with vertical offsets.

\subsubsection{convolutional layers}
In the preliminary study of this paper for horizontal text detection in~\cite{LiaoSBWL17}, we have adopted irregular $1 \times 5$ convolutional filters instead of the standard $3 \times 3$ ones in the text-box layers. This is because that words or text lines in natural images are usually long objects. However, these long convolutional filters are not appropriate for oriented text. Instead, we use $3 \times 5$ convolutional filters for oriented text. These inception-style~\cite{szegedy2015going} filters use rectangular receptive fields, which better fit words with larger aspect ratios. The noise signals that a square-shaped receptive field would bring in are also avoided thanks to these inceptions.

\subsection{Adapted training for arbitrary-oriented text detection}
\label{subsec:training}

\subsubsection{Ground Truth representation}
For the two target box representations described in Section~\ref{subsec:architecture}, we adapt the ground truth representation of oriented text as following:

\paragraph{Quadrilateral}:
For each oriented text ground truth $T$, let $\mathbf{G}_b=(\tilde{x}^b_0, \tilde{y}^b_0, \tilde{w}^b_0, \tilde{h}^b_0)$ be its horizontal rectangle ground truth (\textit{i.e.,} the minimum horizontal rectangle enclosing $T$), where $(\tilde{x}^b_0, \tilde{y}^b_0)$ is the center of $\mathbf{G}_b$, $\tilde{w}^b_0$ and $\tilde{h}^b_0$ are the width and the height of $\mathbf{G}_b$ respectively. This rectangle ground truth can also be denoted as $\mathbf{G}_b = (b_1, b_2, b_3, b_4)$ following Eq.~\eqref{eq:decode-default box}, where $(b_1, b_2, b_3, b_4)$ are the four vertices in clockwise order of $\mathbf{G}_b$ with $b_1$ the top-left one. We use the four vertices of the oriented text ground truth $T$ to represent $T$ in terms of a general quadrilateral denoted by $\mathbf{G}_q = (q_1, q_2, q_3, q_4) = (\tilde{x}^q_1, \tilde{y}^q_1, \tilde{x}^q_2, \tilde{y}^q_2, \tilde{x}^q_3, \tilde{y}^q_3, \tilde{x}^q_4, \tilde{y}^q_4)$. The four vertices $(q_1, q_2, q_3, q_4)$ are also organized in clockwise order such that the sum of Euclidean distances between four point pairs $(b_i, q_i), i = 1, 2, 3, 4$ is minimum. More precisely, let $(q'_1, q'_2, q'_3, q'_4)$ in clockwise order represent the same quadrilateral $\mathbf{G}_q$ with $q'_1$ being the top point (top-left point in case of $\mathbf{G}_q$ being a rectangle). Then the relationship between $q$ and $q'$ is given by:

\begin{align}
\begin{split}
d_{i, \Delta} &= d_E(b_i, q'_{(i+\Delta-1)\%4 + 1}), \Delta =0, 1, 2, 3 \\
\Delta_m &= \arg {\min_\Delta{(d_{1, \Delta}+d_{2, \Delta}+d_{3, \Delta}+d_{4, \Delta})}}, \\
q_i &= q'_{(i+ \Delta_m -1)\%4 + 1},
\end{split}
\end{align}
where $d_E(b,q')$ is the Euclidean distance between two points, and $\Delta_m$ is the shift of points that gives the minimal sum of distance of four pair of corresponding points between $\mathbf{G}_b$ and $\mathbf{G}_q$.
%The principal of determining the first vertex is based on the minimum difficulty of regression.

\paragraph{Rotated rectangle}:
There are several different representations for rotated rectangles. A popular one is given by a horizontal rectangle and a rotated angle, which could be written as $(x, y, w, h, \theta)$. However, due to the bias of the dataset, there is usually an uneven distribution on $\theta$, which may make the model dataset-dependent. To ease this problem, we propose to use another representation proposed in~\cite{r2cnn} for a ground truth rotated rectangle $\mathbf{G}_r$: $\mathbf{G}_r = (\tilde{x}^r_1, \tilde{y}^r_1, \tilde{x}^r_2, \tilde{y}^r_2, \tilde{h}^r)$, where $(\tilde{x}^r_1, \tilde{y}^r_1)$ and $(\tilde{x}^r_2, \tilde{y}^r_2)$ are the first and second vertex of the ground truth (\textit{i.e.,} the first and second vertex of $\mathbf{G}_q$), $\tilde{h}^r$ is the height of the rotated rectangle.

%% The first vertex is set as the top-left one.
%% We just replace the $\mathbf{G}_q$ with $\mathbf{G}_r=(\tilde{x}^r_1, \tilde{y}^r_1, \tilde{x}^r_2, \tilde{y}^r_2, \tilde{h}^r)$ while other setting is the same as the quadrilateral.

\subsubsection{Loss function}
We adopt a loss function similar to the one used in~\cite{liu2015ssd}. More specifically, let $x$ be the match indication matrix. For the $i$-th default box and the $j$-th ground truth, $x_{ij}=1$ means a match following the box overlap between them, otherwise $x_{ij}=0$. Let $c$ be the confidence, $l$ be the predicted location, and $g$ be the ground-truth location. The loss function is defined as:

\begin{equation}
L(x,c,l,g)=\frac{1}{N}(L_{\textrm{conf}}(x,c)+\alpha L_{\textrm{loc}}(x,l,g)),
\end{equation}
where $N$ is the number of default boxes that match ground-truth boxes, and $\alpha$ is set to 0.2 for quick convergence. We adopt the smooth L1 loss~\cite{fast_rcnn} for $L_{\textrm{loc}}$ and a 2-class soft-max loss for $L_{\textrm{conf}}$.

\subsubsection{On-line hard negative mining}
Some textures and signs are very similar to text, which are hard for the network to distinguish. We follow the hard negative mining strategy used in~\cite{liu2015ssd} to suppress them. More precisely, the training on the corresponding dataset is divided into two stages. The ratio between the negatives and positives for the first stage is set to 3:1, and then changed to 6:1 for the second stage. %to further suppress some false positives.

%% We first keep the ratio between the negatives and positives 3:1, and then change it to 6:1 to further suppress the false positives.

\subsubsection{Data Augmentation}
\begin{figure}[!htbp]
\begin{center}
\captionsetup[subfigure]{justification=centering}
    \centering
\subfloat[\label{subfig-1:data-augmentation}]{%
       \includegraphics[width=0.27\textwidth]{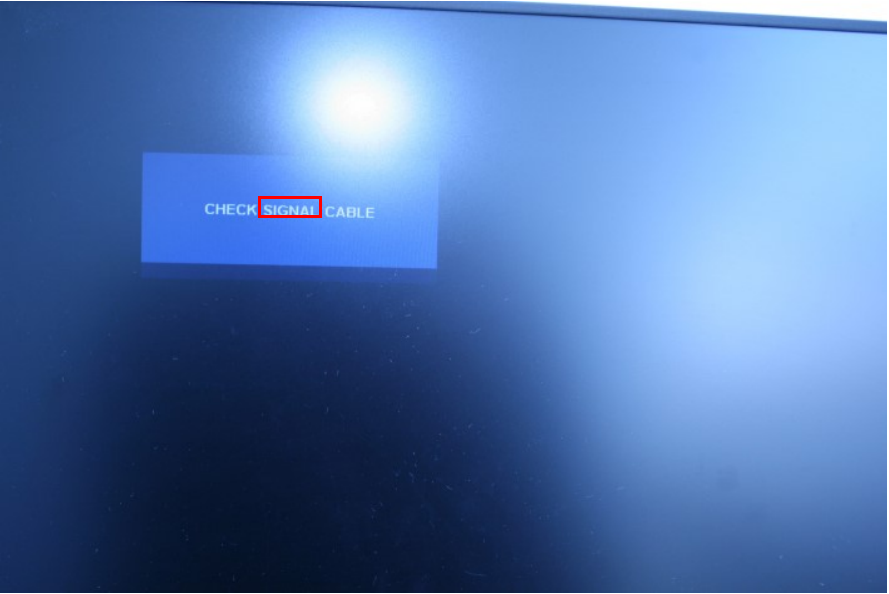}
     }
     \subfloat[\label{subfig-2:data-augmentation}]{%
       \includegraphics[width=0.18\textwidth]{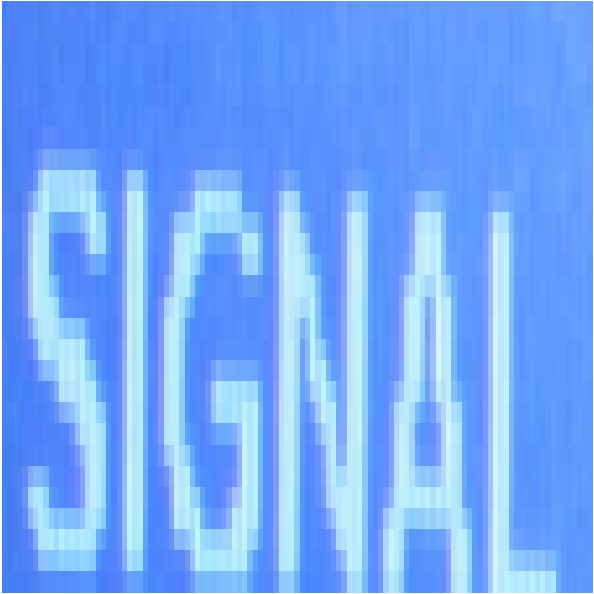}
     }
 	\hfill
\subfloat[\label{subfig-3:data-augmentation}]{%
       \includegraphics[width=0.27\textwidth]{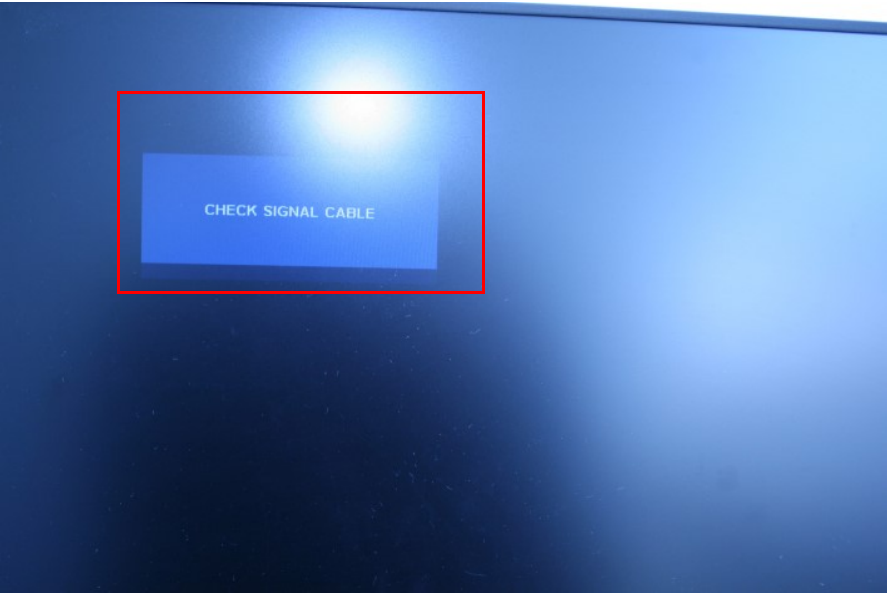}
     }
     \subfloat[\label{subfig-4:data-augmentation}]{%
       \includegraphics[width=0.18\textwidth]{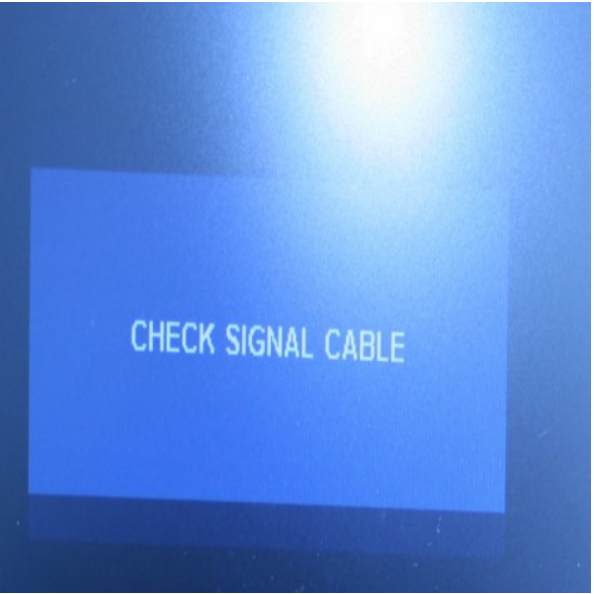}
     }
\end{center}
\caption{Data augmentation by random cropping based on Jaccard overlap (a-b) and object coverage constraints (c-d). Images in (b) and (d) are the corresponding resized crops.}
\label{fig:data-augmentation}
\end{figure}

Similar to many CNN-based vision problems, data augmentation is a classical and necessary way to increase the limited size of a training set. For example, a random crop augmentation based on the minimum Jaccard overlap between crops and ground truths is applied in~\cite{liu2015ssd}. However, this strategy is not appropriate for text which is usually small. This is because that the Jaccard overlap constraint is difficult to satisfy for small objects. As depicted by an example in Fig.~\ref{fig:data-augmentation}(a-b), for a small object, even if the Jaccard overlap is satisfied, the object in the resized image after data augmentation is extremely large that almost covers the whole image. This is not the usual case for text in natural images. To ease this problem, we propose to add a new overlap constraint called object coverage in addition to the Jaccard overlap. For a cropped bounding box $\mathbf{B}$ and a ground truth bounding box $\mathbf{G}$, The Jaccard overlap $\mathbf{J}$ and object coverage $\mathbf{C}$ are defined as follows:
\begin{align}
\begin{split}
  \mathbf{J} &= |\mathbf{B} \cap \mathbf{G}| / |\mathbf{B} \cup \mathbf{G}|, \\
  \mathbf{C} &= |\mathbf{B} \cap \mathbf{G}| / |\mathbf{G}|, \\
\end{split}
\label{eq:overlap strategy}
\end{align}
where $|\cdot|$ denotes the cardinality (\textit{i.e.} area). The random crop strategy based on object coverage $\mathbf{C}$ is more appropriate for small objects such as most text in natural images. An example is given in Fig.~\ref{fig:data-augmentation}(c-d). In this paper, we use both random crop strategies with minimum overlap or coverage thresholds randomly set to 0, 0.1, 0.3, 0.5, 0.7 and 0.9. Note that a threshold set to 0 implies that neither minimum Jaccard overlap nor object coverage constraint is used. Each cropped region is then resized to a fixed size image that feeds into the network. 

\subsubsection{Multi-scale training}
For the sake of training speed, the randomly cropped regions are resized to images of a relatively small size. However, the input images of the proposed TextBoxes++ can have arbitrary size thanks to its fully convolutional architecture. To better handle multi-scale text, we also use larger scale input size for the last several thousand iterations in training phase. The training details are discussed in Section~\ref{Implementation details}.

\subsection{Testing with efficient cascaded non-maximum suppression}
\label{subsec:testing}
Similar to classical methods for object detection, we apply a Non-Maximum Suppression (NMS) during the test period to extract predicted boxes of arbitrary-oriented text. Firstly, we resize the six multi-scale prediction results to the original size of an input image, and fusion these resized results into one dense confidence map. Then the NMS operation is applied on this merged confidence map. Since the NMS operation on quadrilaterals or rotated rectangles is more time-consuming than that on horizontal rectangles, we divide this NMS operation into two steps to accelerate the speed. First, we apply the NMS with a relatively high IOU threshold (\textit{e.g.} 0.5) on the minimum horizontal rectangles containing the predicted quadrilaterals or rotated rectangles. This operation on horizontal rectangles is much less time-consuming and removes many candidate boxes. Then the time-consuming NMS on quadrilaterals or rotated rectangles is applied to a few remaining candidate boxes with a lower IOU threshold (\textit{e.g.} 0.2). The remaining boxes after this second NMS operation are considered as the final detected text boxes. This cascaded non-maximum suppression is much faster than directly applying NMS on the quadrilaterals or rotated rectangles.

\subsection{Word spotting, end-to-end recognition, and detection refining}\label{subsec:word-spotting}
% All benchmarks of scene text detection evaluate the performance of a method by an IOU threshold, which is inherited from object detection. However, text detection is different from object detection because the ultimate purpose of detecting text is to better recognize them. Note that a text recognizer can get the totally different results with the same IOU. As is shown in Figure~\ref{fig:iou}, the detection results in (a), (b) and (c) own the same IOU with the ground truth. However, (a) and (b) lead to a poor recognition result owing to the lacking part of text while (c) is intended to get an accurate recognition result due to better coverage. Thus, we conduct word spotting and end-to-end recognition experiments to further prove the performance of TextBoxes++.

% \begin{figure}[!htbp]
% \begin{center}
% \includegraphics[width=1.0\linewidth]{figures/IOU.pdf}
% \end{center}
% \caption{Different recognition results with the same IOU. The red bounding box is the ground truth and the blue bounding box is the detection result. We can see that all detection results have the equal IOU with the ground truth. However,  detection results on (a) and (b) would get error recognition results while detection result on (c) tends to predict correct recognition result.
% }
% \label{fig:iou}
% \end{figure}

Intuitively, a text recognizer would help to eliminate some false-positive detection results that are unlikely to be meaningful words, \emph{e.g.} repetitive patterns. Particularly, when a lexicon is present, a text recognizer could effectively remove the detected bounding boxes that do not match any of the given words. Following this intuitive idea, we propose to improve the detection results of TextBoxes++ with word spotting and end-to-end recognition. 

\subsubsection{Word spotting and end-to-end recognition}
Word spotting is to localize specific words that are given in a lexicon. End-to-end recognition concerns both detection and recognition. Both tasks can be achieved by simply connecting TextBoxes++ with a text recognizer. We adopt the CRNN model~\cite{shi2015end} as our text recognizer. \revise{CRNN uses CTC~\cite{graves2006connectionist} as its output layer, which estimates a sequence probability conditioned on the input image $I$ denoted as $p(\textbf{w} | I)$, where $\textbf{w}$ represents a character sequence output. If no lexicon is given, $\textbf{w}$ is considered as the recognized word, and the probability $p(\textbf{w} | I)$ measures the compatibility of an image to that particular word $\textbf{w}$. CRNN also supports the use of the lexicon. For a given lexicon ${\cal W}$, CRNN outputs the probability that measures how the input image $I$ matches each word $w \in {\cal W}$. We define the recognition score $s_r$ in the following:}
% Note that $\textbf{w}$ can either be an output of the model (without lexicon) or a word in a given lexicon. We consider the probability $p(\textbf{w} | I)$ as a matching score, which measures the compatibility of an image to a particular word. The recognition score is then defined as the matching score (without lexicon) or the maximum score among all words in a given lexicon:
%
\begin{equation}
\revise{s_r=}\revise{\left\{\begin{matrix}
p(\textbf{w} | I), & \textrm{Without lexicon}  \\ 
\max\limits_{\textbf{w} \in {\cal W} } p(\textbf{w} | I), & \textrm{With lexicon } {\cal W}
\end{matrix}\right.}
%   s = \max_{\textbf{w} \in {\cal W} } p(\textbf{w} | I), if W exists\\
%   s = p(\textbf{w} | I), if W not exists
\label{eq:ctc-score}
\end{equation}
\revise{Note that the use of lexicon is not a necessary in the proposed method. We only use lexicons for fair comparisons with other methods.}

\subsubsection{Refining detection with recognition}
We propose to refine detection with recognition by integrating the recognition score $s_r$ to the original detection $s_d$ score. In practice, the value of recognition score is generally not comparable to the value of detection score. For example, the threshold of the detection score $s_d$ is usually set to 0.6, and the threshold of recognition score $s_r$ is usually set to 0.005. A trivial combination of these two scores would lead to a severe bias of detection score. In this paper, we propose to define the novel score $S$ as following:
\begin{equation}
  S = \frac{2 \times e^{(s_d+s_r)}}{e^{s_d}+e^{s_r}}.
\label{eq:synthetic-score}
\end{equation}

There are two motivations in Eq.~\eqref{eq:synthetic-score}. First, we use the exponential function to make the two score values comparable. Then, a harmonic mean is adopted to get the final combined score. This combined score $S$ is more convenient than applying a grid search on two scores, respectively.

%proposed synthetic score gently combines the detection score and the recognition score, which is more convenient than applying grid search on two scores.

\section{Experiments} \label{sec: experimental results}

\begin{figure}[!htbp]
\begin{center}
\captionsetup[subfigure]{justification=centering}
    \centering
\subfloat[\label{subfig-1:iou}]{%
       \includegraphics[width=0.15\textwidth]{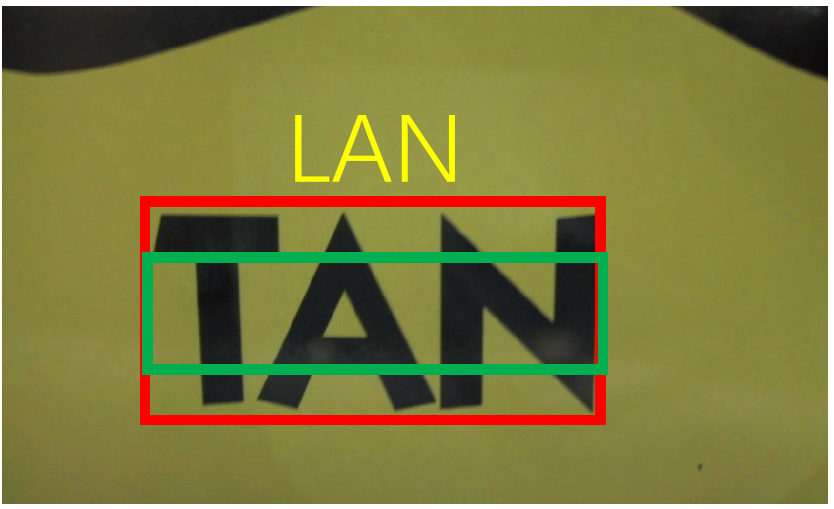}
     }
\subfloat[\label{subfig-2:iou}]{%
       \includegraphics[width=0.15\textwidth]{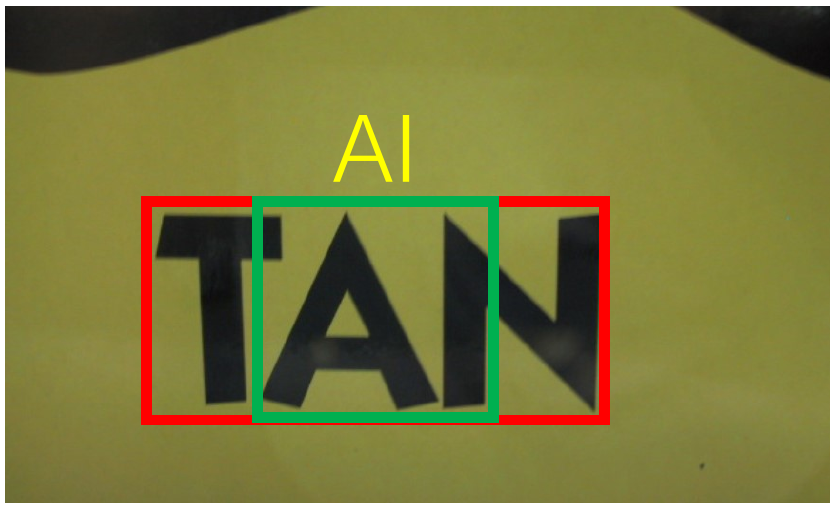}
     }
\subfloat[\label{subfig-3:iou}]{%
       \includegraphics[width=0.15\textwidth]{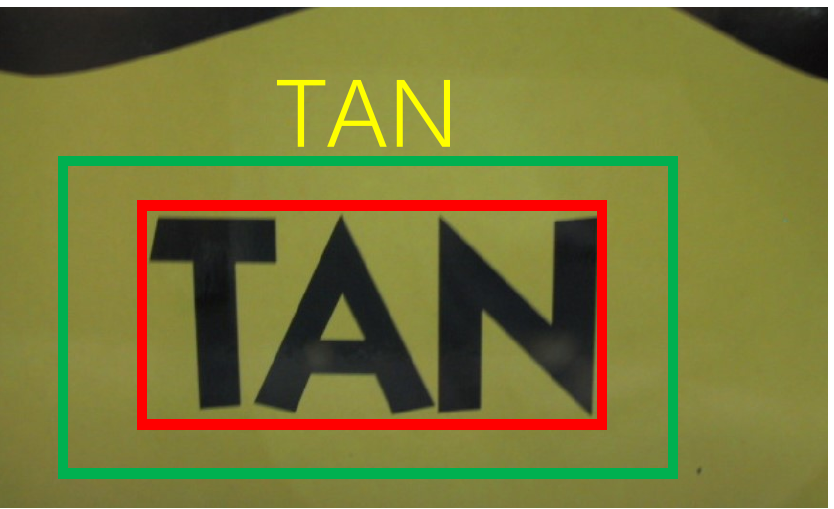}
     }
\end{center}
\caption{An example of recognition results (yellow text) with different text detection results having the same IOU. The red (\textit{resp.} green) bounding box is the ground truth (\textit{resp.} detection result).}
\label{fig:iou}
\end{figure}

Inherited from object detection, all existing scene text detection benchmarks rely on an IOU threshold to evaluate the performance of text detectors. However, text detection is quite different from object detection because the ultimate purpose of detecting text is text recognition. A text recognizer may yield totally different results with the same IOU. For example, the three detection results in Fig.~\ref{fig:iou}(a-c) have the same IOU. However, the detection results in Fig.~\ref{fig:iou}(a) and Fig.~\ref{fig:iou}(b) fail to correctly recognize the underlying text due to the lack of text parts. The detection results in Fig.~\ref{fig:iou}(c) tends to give an accurate recognition result thanks to the full text coverage. Thus, in addition to classical text detection benchmarks, we also conduct word spotting and end-to-end recognition experiments using the text detection results to further demonstrate the performance of TextBoxes++.

\subsection{Datasets \& evaluation protocol} 
\label{subsec:datasets}

The proposed TextBoxes++ detects arbitrary-oriented text. We have tested its performance on two datasets including oriented text: ICDAR 2015 Incidental Text (IC15) dataset~\cite{icdar/KaratzasGNGBIMN15} and {COCO-Text} dataset~\cite{coco-text/VeitMNMB16}. To further demonstrate the versatility of TextBoxes++, we have conducted experiments on two popular horizontal text datasets: ICDAR 2013 (IC13) dataset~\cite{karatzas2013icdar} and Street View Text (SVT) dataset~\cite{wang2010word}. Besides these benchmark datasets, the SynthText dataset~\cite{gupta2016synthetic} is also used to pre-train our model. A short description of all these concerned datasets is given in the following (see the corresponding references for more details).

\noindent\textbf{SynthText}: The SynthText dataset~\cite{gupta2016synthetic} contains 800k synthesized text images, created via blending rendered words with natural images. As the location and transform of text are carefully chosen with a learning algorithm, the synthesized images look realistic. This dataset is used for pre-training our model.

\noindent\textbf{ICDAR 2015 Incidental Text (IC15)}: The ICDAR 2015 Incidental Text dataset~\cite{icdar/KaratzasGNGBIMN15} issues from the Challenge 4 of the ICDAR 2015 Robust Reading Competition. The dataset is composed of 1000 training images and 500 testing images, which are captured by Google glasses with relatively low resolutions. Each image may contain multi-oriented text. Annotations are provided in terms of word bounding boxes. This dataset also provides 3 lexicons of different sizes for word spotting and end-to-end recognition challenge: 1) strong lexicon which gives 100 words as an individual lexicon for each test image; 2) weakly lexicon containing hundreds of words for the whole test set; 3) generic lexicon with 90k words.

\noindent\textbf{COCO-Text}: The COCO-Text dataset~\cite{coco-text/VeitMNMB16} is currently the largest dataset for scene text detection and recognition. It contains 43686 training images and 20000 images for validation/testing. The COCO-Text dataset is very challenging since text in this dataset are in arbitrary orientations. This difficulty also holds for annotations which are not as accurate as the other tested datasets in this paper. Therefore, even though this dataset provides oriented annotations, its standard evaluation protocol still relies on horizontal bounding rectangles. For TextBoxes++, we make use of both the annotations in terms of horizontal bounding rectangles and the quadrilateral annotations to train our model. For evaluation, we follow the standard protocol based on horizontal bounding rectangles.

\noindent\textbf{ICDAR 2013 (IC13)}: The ICDAR 2013 dataset~\cite{karatzas2013icdar} consists of 229 training images and 233 testing images in different resolutions. This dataset contains only horizontal or nearly horizontal text. The lexicon setting for this dataset is the same as the IC15 dataset described before.

%% The ICDAR 2013 dataset gives 3 lexicons of different sizes for the task of word spotting and end-to-end recognition. For each test image, it gives 100 words as a lexicon, which is called a strong lexicon. For the whole test set, it gives a lexicon containing hundreds of words, which is called a weakly lexicon. It also gives a generic lexicon which contains 90k words. The lexicon setting is the same as the ICDAR 2013.

\noindent\textbf{Street View Text (SVT)}: The SVT dataset~\cite{wang2010word} is more challenging than previous ICDAR 2013 dataset due to lower resolutions of images. There are 100 training images and 250 testing images in the SVT dataset. The images have only horizontal or nearly horizontal text. A lexicon containing 50 words is also provided for each image. Note that not all the text in the dataset are labeled. As a result, this dataset is only used for word spotting evaluation.

\noindent\textbf{Evaluation Protocols}: The classical evaluation protocols for text detection, word spotting, and end-to-end recognition all rely on $precision$ (P), $recall$ (R), and $f$-$measure$ (F). They are given by:
\begin{align}
\begin{split}
P &= \frac{TP}{TP+FP}\\
R &= \frac{TP}{TP+FN}\\ 
F &= 2 \times \frac{P \times R}{P + R}
\end{split}
\end{align}
where TP, FP, and FN is the number of hit boxes, incorrectly identified boxes, and missed boxes, respectively. For text detection, a detected box $b$ is considered as a hit box if the IOU between $b$ and a ground truth box is larger than a given threshold (generally set to 0.5). The hit boxes in word spotting and end-to-end recognition require not only the same IOU restriction but also correct recognition results. Since there is a trade-off between precision and recall, $f$-$measure$ is the most used measurement for performance assessment.

\subsection{Implementation details} \label{Implementation details}
% \textbf{TextBoxes} is trained with 300*300 images using stochastic gradient descent (SGD). Momentum and weight decay are set to $0.9$ and $5\times 10^{-4}$ respectively. Learning rate is initially set to $10^{-3}$, and decayed to $10^{-4}$ after 40k training iterations. On all the datasets except SVT, we first train TextBoxes on SynthText for 50k iterations, then fine-tune it on ICDAR 2013 training dataset for 2k iterations. On SVT, the fine-tuning is performed on the SVT training dataset. All training images are augmented on-line with random crop and flip, following the scheme in~\cite{liu2015ssd}.
\begin{table}[!htbp]
\centering
\caption{Implementation details. ``lr'' is short for learning rate. The size of input image is denoted as ``size''.``nr'' refers to the negative ratio in hard negative mining. ``\#iter'' stands for the number of training iterations.}
\label{table: implementation details}
\begin{tabular}{|c|c|c|c|c|c|c|c|}
\hline
Dataset   & \multicolumn{3}{c|}{All datasets} & IC15 & COCO-Text & SVT    & IC13 \\ \hline
Settings  & lr             & size     & nr    & \#iter                     & \#iter     & \#iter & \#iter     \\ \hline
Pre-train & $10^{-4}$     & 384      & 3     & 60k                        & 60k        & 60k    & 60k        \\ \hline
Stage 1   & $10^{-4}$      & 384      & 3     & 8k                         & 20k        & 2k     & 2k         \\ \hline
Stage 2   & $10^{-5}$      & 768      & 6     & 4k                         & 30k        & 8k     & 8k         \\ \hline
\end{tabular}
\end{table}

% \textcolor{red}{Prepare a table to explain these settings. Then only summarizing text are needed. For example, the whole training process is composed of three stages: 1) pre-training on SynthText; 2) learning rate and number of iterations in second stage; 3) the same setting details as 2nd stage. Multi-scale settings in Method also find place here.}
\textbf{TextBoxes++} is trained with Adam~\cite{adam}. The whole training process is composed of three stages as shown in Table.~\ref{table: implementation details}. Firstly, we pre-train TextBoxes++ on SynthText dataset for all tested datasets. Then the training process is continued on the corresponding training images of each dataset. Finally, we continue this training with a smaller learning rate and a larger negative ratio. Note also that at the last training stage, a larger input image size is used to achieve better detections of multi-scale text. The number of iterations for the pre-training step is fixed at 60k for all tested datasets. However, this number differs in the second and third training stage which are conducted on each dataset's own training images. This difference is decided by the different dataset size. %For simplification, We fix the quadrilateral branch for horizontal dataset such as ICDAR 2013 and SVT.

% \textbf{TextBoxes++} is trained with 384*384 images using Adam\cite{adam}.
% All models of TextBoxes++ are pre-trained on SynthText for 60k iterations, with base learning rate setting to $10^{-4}$. \\
% For ICDAR 2015 Incidental Dataset, we first use%using
% $384*384$ to train 8k with the base learning rate setting to $10^{-4}$, and then %we
% enlarge the input size to 768*768, and decay the learning rate to $10^{-5}$ at the same time for the rest 4k iterations. The negative ratio of hard negative mining is set to 3 except for the last 4k iterations, during which the negative ratio is set to 6 to further suppress the false positives.\\
% COCO-Text is much larger than ICDAR 2015 Incidental Text. We follow the same training scheme of ICDAR 2015 while train more iterations for COCO-Text. First, we train the network for 20k iterations with learning rate setting to $10^{-4}$, during which the image size is set to 384*384. Then, we enlarge the input size to 768*768 and decay the learning rate to $10^{-5}$ for the last 30k iterations.
% SVT and ICDAR 2013 dataset are trained following the training scheme of TextBoxes~\cite{LiaoSBWL17} with training tricks above. Note that multi-scale training strategy is also involved in the training.

\textbf{Text recognition} is performed with a pre-trained CRNN model~\cite{shi2015end}, which is implemented and released by the authors\footnote{https://github.com/bgshih/crnn}.

All the experiments presented in this paper are carried out on a PC equipped with a single Titan Xp GPU. The whole training process (including the pre-training time on SynthText dataset) takes about 2 days on ICDAR 2015 Incidental Text dataset, which is currently the most tested dataset.

\begin{figure}[!ht]
\begin{center}
\captionsetup[subfigure]{justification=centering}
    \centering
\begin{minipage}{.20\textwidth}
\subfloat[\label{subfig-1:ground-truth-resize}]{%
       \includegraphics[width=1.0\textwidth]{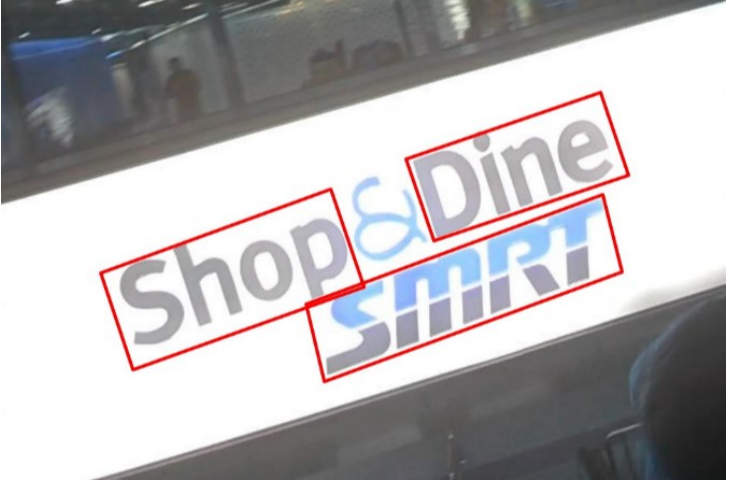}
     }
\hfill
\subfloat[\label{subfig-2:ground-truth-resize}]{%
       \includegraphics[width=1.0\textwidth]{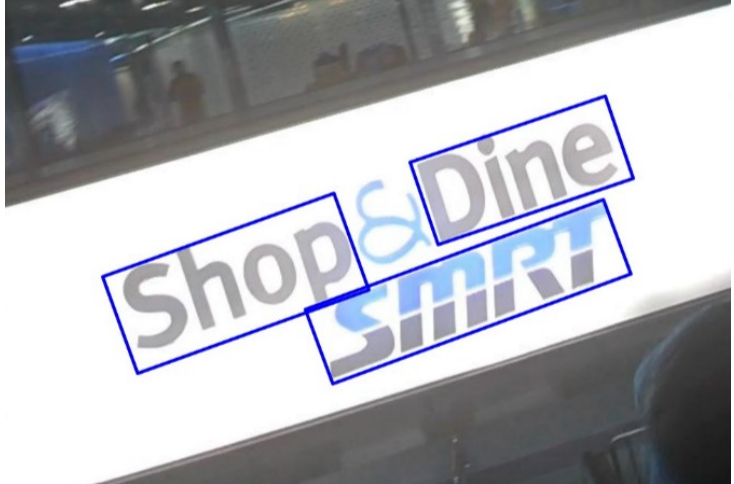}
     }
\end{minipage}%
% \vspace{0.15\textwidth}
\begin{minipage}{.26\textwidth}
\hspace{0.02\textwidth}
\subfloat[\label{subfig-3:ground-truth-resize}]{%
       \includegraphics[width=0.45\textwidth]{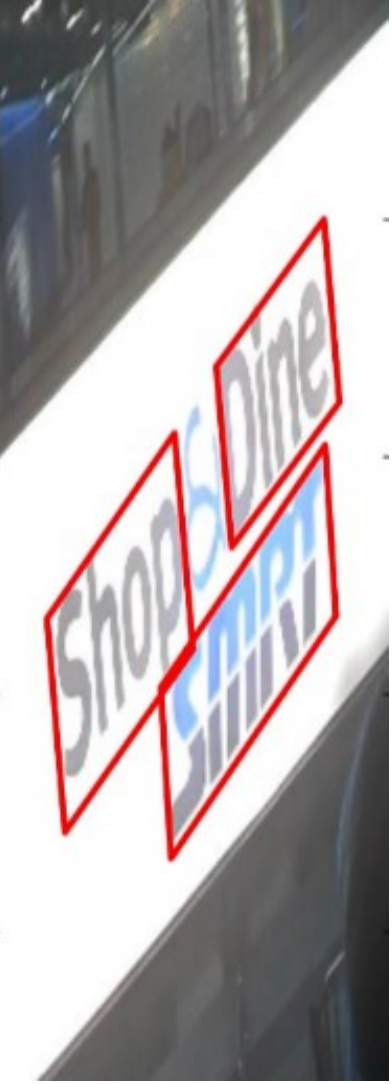}
     }
% \hspace{0.01\textwidth}
\subfloat[\label{subfig-4:ground-truth-resize}]{%
       \includegraphics[width=0.45\textwidth]{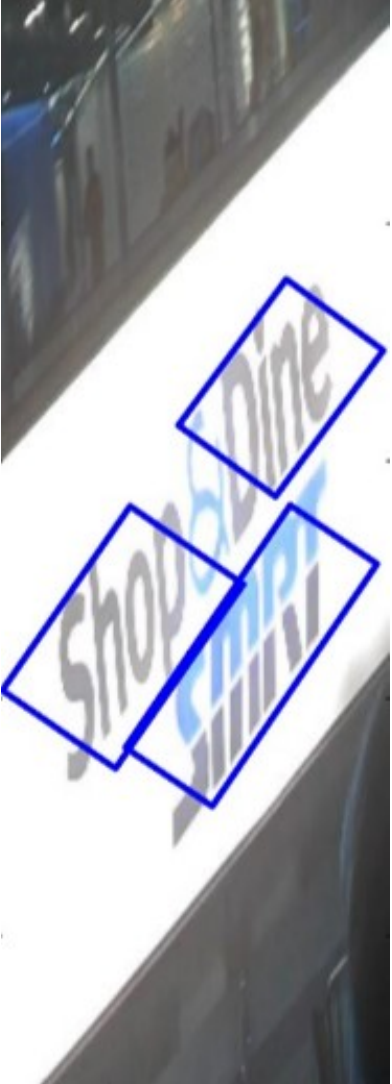}
     }
\end{minipage}%  
\end{center}
\caption{Ground truth representations in terms of quadrilaterals in red and rotated rectangles in blue. The underlying image in (c) and (d) is resized from the original image shown both in (a) and (b).}
\label{fig:ground-truth-resize}
\end{figure}

\subsection{Quadrilateral VS Rotated Rectangle} \label{sec:quadrilateral vs rotated rectangle}
\begin{table}[!tbp]
\centering
\caption{Performance comparisons on ICDAR 2015 Incidental Text dataset with different IOU threshold settings between four variants of TextBoxes++, which use different bounding box representation and different input scales. ``RR'' stands for the rotated rectangle, and ``Quad'' represents the quadrilateral. ``MS'' is short for using multi-scale inputs.
}
\label{polygon_vs_rbox}
\begin{tabular}{|c|c|c|c|c|c|c|}
\hline
\multirow{2}{*}{Method} & \multicolumn{3}{c|}{IOU\_threshold=0.5} & \multicolumn{3}{c|}{IOU\_threshold=0.7} \\ \cline{2-7} 
                        & R           & P           & F           & R           & P           & F           \\ \hline
RR         &  0.764      & 0.822       & 0.792       & 0.613            & 0.574            & 0.593            \\ \hline
Quad       &  0.767      & 0.872       & 0.817       & 0.577            & 0.676            & 0.623            \\ \hline
RR\_MS     &  0.766      & 0.875       & 0.817       & 0.569            & 0.623            & 0.594            \\ \hline
Quad\_MS   &  \bf{0.785}      & \bf{0.878}       & \bf{0.829}       & \bf{0.617}            & \bf{0.690}            & \bf{0.651}            \\ \hline
\end{tabular}
\end{table}

The rotated rectangle is an approximate simplification of the quadrilateral, which is more flexible in representing arbitrary-oriented text bounding box. Although both representations may equally suit for normal text fonts and styles, the quadrilateral representation may adapt better to resized images. An example is given in Fig.~\ref{fig:ground-truth-resize}, where the bounding boxes represented by quadrilaterals and rotated rectangles in the original image (see Fig.~\ref{fig:ground-truth-resize}(a-b)) are almost the same. However, as shown in Fig.~\ref{fig:ground-truth-resize}(c-d), the rotated rectangles match less well the text than quadrilaterals in the resized image. This is because that a rotated rectangle generally become a parallelogram when resized directly, which leads to a small deviation when trying to keep it as a rotated rectangle. In this sense, TextBoxes++ with quadrilateral representation would be more accurate than its variant using rotated rectangles.

We have conducted experiments on the widely tested ICDAR 2015 Incidental Text dataset to compare these two variants of TextBoxes++; The quantitative comparison is given in Table~\ref{polygon_vs_rbox}. Following the standard text detection evaluation, the TextBoxes++ using quadrilateral representation significantly outperforms the version using rotated rectangles with at least 2.5 percents improvements. Note also that using multi-scale inputs would improve both versions of TextBoxes++. The quadrilateral version still performs better especially when the IOU threshold for matching evaluation is set to 0.7. Besides, under such a high IOU threshold setting, the difference between quadrilateral version and rotated rectangle version with multi-scale inputs is more significant. This is because a much more accurate text detector is expected for a high IOU threshold setting. This confirms that quadrilateral representation is more accurate than the rotated rectangle for TextBoxes++. Consequently, we choose the TextBoxes++ using the quadrilateral representation for the rest experiments in this paper and denote it simply as TextBoxes++ when no ambiguity is present. TextBoxes++\_MS stands for this version of TextBoxes++ with multi-scale inputs.

%% More over, when IOU threshold is set to 0.7, TextBoxes++\_Quad\_MS is 2.8 percents higher than TextBoxes++\_Quad while TextBoxes++\_RR\_MS only increases 0.1 percent compared with TextBoxes++\_RR, which means that the quadrilateral version is much more robust with different scales.

%% In conclusion,  the bounding boxes of quadrilateral version has two advantages. On the one hand, it is more accurate than the rotated rectangle version, which is especially obvious when IOU threshold is set to 0.7. On the other hand, it is much more robust with different input scales, which reflects in the performance gain when applying multi-scale testing.

\subsection{Text localization}

\begin{figure*}[!ht]
\begin{center}
\captionsetup[subfigure]{justification=centering}
    \centering
\subfloat[Results given by Zhang et al.~\cite{Zhang_2016_CVPR}\label{subfig-1:localization-visualization}]{%
       \includegraphics[width=0.99\textwidth]{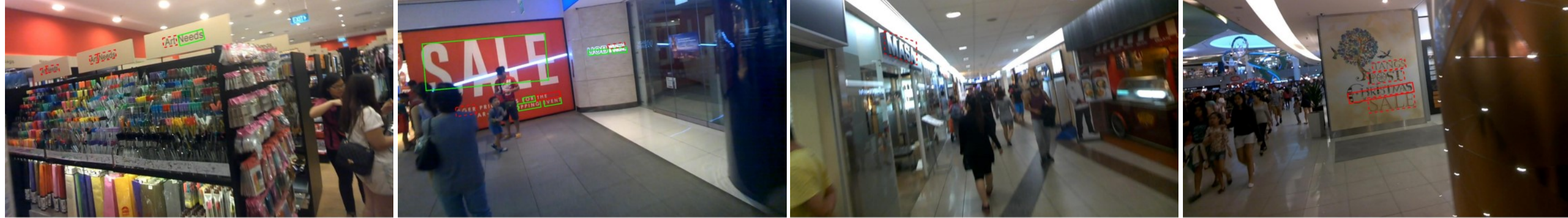}
     }
\hfill
\subfloat[Results using Shi et al.~\cite{corr/ShiBB17}\label{subfig-2:localization-visualization}]{%
       \includegraphics[width=0.99\textwidth]{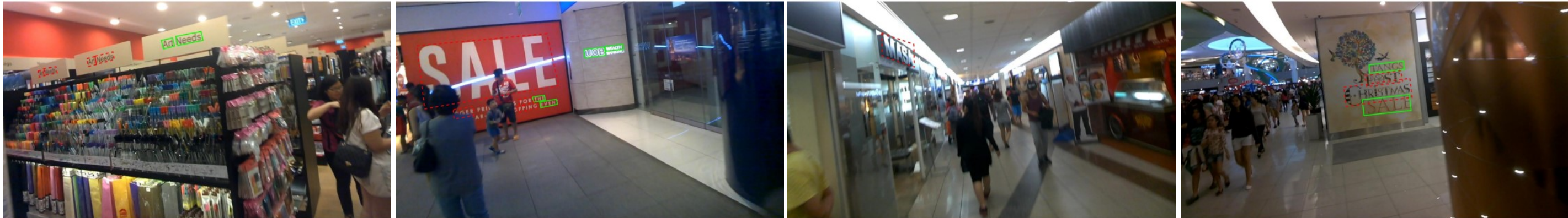}
     }
\hfill
\subfloat[TextBoxes++ results\label{subfig-3:localization-visualization}]{%
       \includegraphics[width=0.99\textwidth]{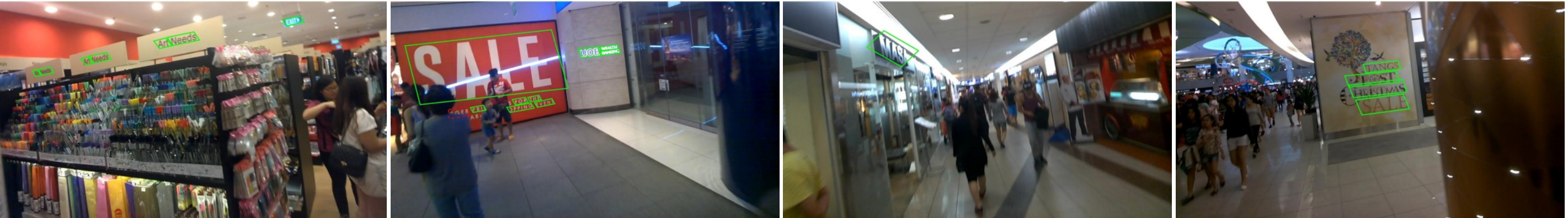}
     }
\end{center}
% \vspace{-5mm}
\caption{Qualitative comparisons of text detection results on some ICDAR 2015 Incidental text images. %The results of the first row are from Zhang et al.~\cite{Zhang_2016_CVPR} and the results of the second row are from Shi et al.~\cite{corr/ShiBB17}. The results of TextBoxes++ are placed on the last row.
Green bounding boxes: correct detections; Red solid boxes: false detections; Red dashed boxes: missed ground truths.
}
\label{fig:localization-visualization}
% \vspace{-2mm}
\end{figure*}

\begin{table}[!tbp]
\centering
\caption{Text localization results on ICDAR 2015 Incidental Text dataset.}
\label{icdar15 text detection}
\begin{tabular}{|c|c|c|c|}
\hline
Methods      & recall & precision & f-measure \\ \hline
CNN MSER~\cite{icdar/KaratzasGNGBIMN15}     & 0.34   & 0.35      & 0.35      \\ \hline
% Deep2Text-MO~\cite{yin2014robust,yin2015multi} & 0.32   & 0.50      & 0.39      \\ \hline
AJOU~\cite{AJOU}         & 0.47   & 0.47      & 0.47      \\ \hline
NJU~\cite{icdar/KaratzasGNGBIMN15}          & 0.36   & 0.70      & 0.48      \\ \hline
StradVision1~\cite{icdar/KaratzasGNGBIMN15} & 0.46   & 0.53      & 0.50      \\ \hline
StradVision2~\cite{icdar/KaratzasGNGBIMN15} & 0.37   & 0.77      & 0.50      \\ \hline
Zhang et al.~\cite{Zhang_2016_CVPR} & 0.43   & 0.71      & 0.54      \\ \hline
Tian et al.~\cite{eccv/TianHHH016}  & 0.52   & 0.74      & 0.61      \\ \hline
Yao et al.~\cite{corr/YaoBSZZC16}   & 0.59   & 0.72      & 0.65      \\ \hline
Liu et al.~\cite{LiuJ17b}   & 0.682   & 0.732      & 0.706      \\ \hline
Shi et al.~\cite{corr/ShiBB17}   & 0.768   & 0.731      & 0.750      \\ \hline
EAST PVANET2x. RBOX~\cite{corr/EAST}   & 0.735   & 0.836      & 0.782      \\ \hline
EAST PVANET2x RBOX MS~\cite{corr/EAST}   & 0.783   & 0.833      & 0.807      \\ \hline
%TextBoxes++\_RR  & 0.764 & 0.822 & 0.792        \\ \hline
TextBoxes++  & 0.767  & 0.872  &  \bf{0.817}  \\ \hline
%TextBoxes++\_RR\_MS & 0.766   & 0.875   & 0.817  \\ \hline
TextBoxes++\_MS & \bf{0.785}  & 0.\bf{878}  & \bf{0.829}  \\ \hline
\end{tabular}
\end{table}

\begin{table}[!tbp]
\centering
\caption{Text localization results on COCO-Text dataset.}
\label{coco-text detection}
\begin{tabular}{|c|c|c|c|}
\hline
Methods    & recall & precision & f-measure \\ \hline
Baseline A~\cite{coco-text/VeitMNMB16} & 0.233  & 0.8378    & 0.3648    \\ \hline
Baseline B~\cite{coco-text/VeitMNMB16} & 0.107  & 0.8973    & 0.1914    \\ \hline
Baseline C~\cite{coco-text/VeitMNMB16} & 0.047  & 0.1856    & 0.0747    \\ \hline
Yao et al.~\cite{corr/YaoBSZZC16} & 0.271  & 0.4323    & 0.3331    \\ \hline
Zhou et al.~\cite{corr/EAST} & 0.324 & 0.5039 & 0.3945 \\ \hline
TextBoxes++ & 0.5600  & 0.5582  & 0.5591   \\ \hline
TextBoxes++\_MS  & \bf{0.5670}  & \bf{0.6087}  & \bf{0.5872}  \\ \hline
\end{tabular}
\end{table}

\begin{table}[!htbp]
\centering
\caption{Text localization on ICDAR 2013 dataset. P, R, and F refer to precision, recall and f-measure, respectively.}
\label{tab:text-localization}
\begin{tabular}{|c|c|c|c|c|c|c|}
\hline
Evaluation protocol                            & \multicolumn{3}{c|}{IC13 Eval}                      & \multicolumn{3}{c|}{DetEval}                    \\ \hline
Methods                                        & R               & P               & F               & R               & P           & F               \\ \hline
% Liang et al.~\cite{tip/LiangSLT15}        & 0.66            & 0.76            & 0.70            & --              & --          & --              \\ \hline
% Wu et al.~\cite{wu2015new}        & 0.70            & 0.76            & 0.73            & --              & --          & --              \\ \hline
fasttext~\cite{busta2015fastext}        & 0.69            & 0.84            & 0.77            & --              & --          & --              \\ \hline
MMser~\cite{zamberletti2014text}        & 0.70            & 0.86            & 0.77            & --              & --          & --              \\ \hline
Lu et al.~\cite{lu2015scene}        & 0.70            & 0.89            & 0.78            & --              & --          & --              \\ \hline
TextFlow~\cite{tian2015text}            & 0.76            & 0.85            & 0.80            & --              & --          & --              \\ \hline
He et al.~\cite{he2016aggregating}            & 0.76            & 0.85            & 0.80            & --              & --          & --              \\ \hline
He et al.~\cite{he2016text}            & 0.73            & \textbf{0.93}            & 0.82            & --              & --          & --              \\ \hline
FCRNall+filts~\cite{gupta2016synthetic} & --              & --              & --              & 0.76            & 0.92        & 0.83            \\ \hline
FCN~\cite{Zhang_2016_CVPR}            & 0.78            & 0.88            & 0.83            & --              & --          & --              \\ \hline
Tian et al~\cite{tian2017natural}            & 0.84            & 0.84            & 0.84            & --              & --          & --              \\ \hline
Qin et al.~\cite{qin2016fast}            & 0.79            & 0.89            & 0.83            & --              & --          & --              \\ \hline
Shi et al.~\cite{corr/ShiBB17}          & --              & --              & --              & 0.83            & 0.88        & 0.85            \\ \hline
Tian et al.~\cite{eccv/TianHHH016}      & --              & --              & --              & 0.83            & \textbf{0.93} & 0.88            \\ \hline
Tang et al.~\cite{tip/TangW17}     & 0.87      & 0.92 & \textbf{0.90}    & --              & --              & --                          \\ \hline
SSD~\cite{liu2015ssd}                   & 0.60            & 0.80            & 0.68            & 0.60            & 0.80        & 0.69            \\ \hline
TextBoxes~\cite{LiaoSBWL17}                                      & 0.74            & 0.86            & 0.80            & 0.74            & 0.88        & 0.81            \\ \hline
TextBoxes MS~\cite{LiaoSBWL17}                                   & 0.83            & 0.88            & 0.85            & 0.83            & 0.89        & 0.86            \\ \hline
TextBoxes++                                    & 0.74            & 0.86            & 0.80            & 0.74            & 0.88        & 0.81            \\ \hline
TextBoxes++\_MS                                 & \textbf{0.84} & \textbf{0.91} & 0.88 & \textbf{0.86} & 0.92        & \textbf{0.89} \\ \hline
\end{tabular}
\end{table}

\subsubsection{Performance}
We have first evaluated the proposed TextBoxes++ on two popular oriented text datasets to assess its ability of handling arbitrary-oriented text in natural images. To further validate the versatility of TextBoxes++, we have also tested on two widely used horizontal text datasets.

\textbf{Oriented Text Dataset}: The first tested oriented text dataset is the widely used ICDAR 2015 Incidental text dataset. Some qualitative comparisons are illustrated in Fig.~\ref{fig:localization-visualization}. As depicted in this figure, TextBoxes++ is more robust than the competing methods in detecting oriented text and text of a variety of scales. Quantitative results following the standard evaluation protocol is given in Table.~\ref{icdar15 text detection}. TextBoxes++ with single input scale outperforms all the state-of-the-art results. More specifically, TextBoxes++ improves the state-of-the-method~\cite{corr/EAST} by 3.5 percents when single scale inputs are used in both methods. Furthermore, the single scale version of TextBoxes++ is still 1.0 percent than the multi-scale version of~\cite{corr/EAST}. Note that a better performance is achieved for TextBoxes++ using multi-scale inputs.

TextBoxes++ also significantly exceeds the state-of-the-art methods on COCO-Text dataset with its latest annotations v1.4. As depicted in Table.~\ref{coco-text detection}, TextBoxes++ outperforms the competing methods by at least 16 percents with single scale input. Furthermore, the performance of TextBoxes++ is boosted by 2.81 percents when multi-scale inputs are used.

\textbf{Horizontal Text Dataset}: We have also evaluated TextBoxes++ on ICDAR 2013 dataset,  one of the most popular horizontal text dataset. The comparison with some state-of-the-art methods is depicted in Table.~\ref{tab:text-localization}. Note that there are many methods evaluated on this dataset, but only some of the best results are shown. TextBoxes++ achieves at least 1.0 percent improvement over other methods except for~\cite{tip/TangW17} on this dataset. However, Tang et al.~\cite{tip/TangW17} use a cascaded architecture which contains two networks, taking 1.36 seconds per image. Moreover, it can only detect horizontal text. We have also compared TextBoxes++ with one state-of-the-art general object detector SSD~\cite{liu2015ssd}, which is the most related method. The same training procedures of TextBoxes++ is used to train SSD for this comparison. As reported in Table.~\ref{tab:text-localization}, such a straightforward adaption of SSD for text detection does not perform as well as the state-of-the-art methods. In particular, we have observed that SSD fails to correctly detect the words with large aspect ratios. TextBoxes++ performs much better thanks to the proposed text-box layers which are specifically designed to overcome the length variation of words. \revise{Compared with TextBoxes~\cite{LiaoSBWL17}, TextBoxes++ achieves almost the same performance with a single scale, and better performance is achieved with multi-scales, owing to the multi-scale training strategy adopted in TextBoxes++. However, note that the experiment on this dataset is to verify that TextBoxes++, although dedicated to arbitrary-oriented text detection, has no performance loss compared with the preliminary study TextBoxes, which is specifically designed for horizontal text detection. }

\begin{figure*}[!htbp]
\begin{center}
\captionsetup[subfigure]{justification=centering}
    \centering
\subfloat[Some results on ICDAR 2013 images\label{subfig-1:spotting-visualization}]{%
       \includegraphics[width=1.0\textwidth]{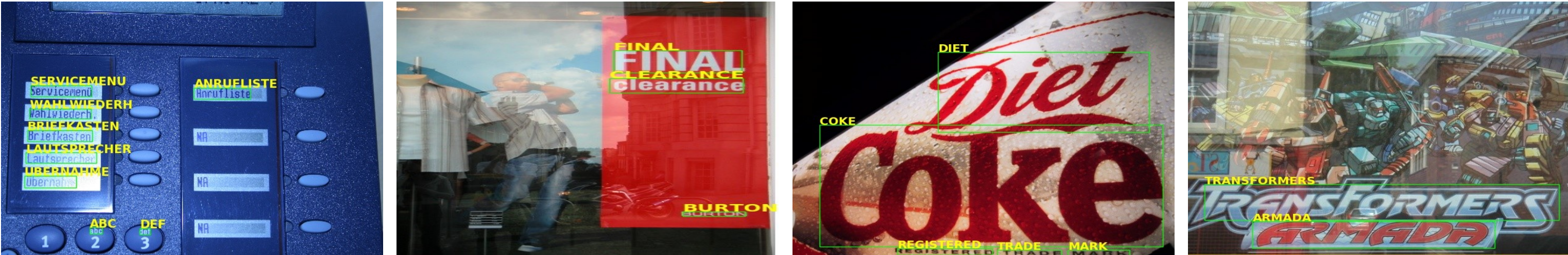}
     }
\hfill
\subfloat[Some results on ICDAR 2015 Incidental text images\label{subfig-2:spotting-visualization}]{%
       \includegraphics[width=0.99\textwidth]{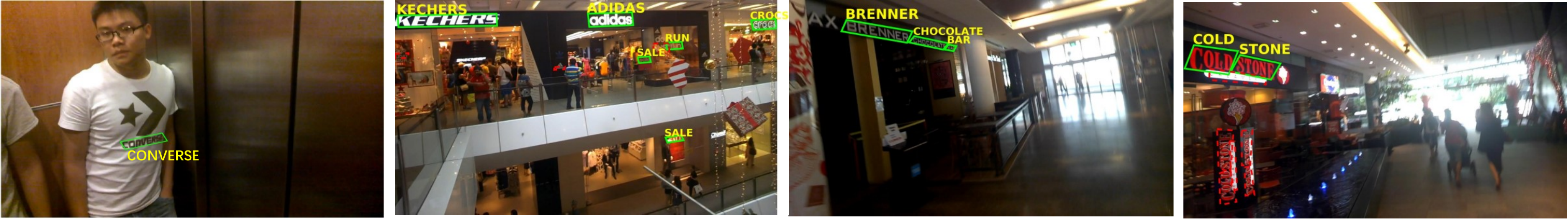}
     }
\end{center}
\caption{Some examples of end-to-end recognition results represented by yellow words. Note that following the evaluation protocol, Words less than 3 letters are ignored. The box colors have the same meaning as Fig.~\ref{fig:localization-visualization}.}
\label{fig:spotting-visualization}
\end{figure*}

% \begin{table}[!htbp]
% \centering
% \caption{Runtime and performance comparison on ICDAR 2015 Incidental Text dataset. ``F'' is short for F-measure. See corresponding text for detailed discussions.}
% \label{table: speed compare}
% \begin{tabular}{|c|c|c|c|c|}
% \hline
% Method               & Res.    & Device  & FPS   & F \\ \hline
% Zhang et al.~\cite{Zhang_2016_CVPR}         & MS*     & Titan X & 0.476 & 0.54      \\ \hline
% Tian et al.~\cite{eccv/TianHHH016}         & ss-600* & GPU     & 7.14  & 0.61      \\ \hline
% Yao et al. ~\cite{corr/YaoBSZZC16}          & 480p    & K40m    & 1.61  & 0.65      \\ \hline
% EAST PVANET~\cite{corr/EAST}   & 720p    & Titan X & \textbf{16.8}  & 0.757    \\ \hline
% EAST PVANET2x~\cite{corr/EAST} & 720p    & Titan X & 13.2  & 0.782     \\ \hline
% EAST VGG16~\cite{corr/EAST}    & 720p    & Titan X & 6.52  & 0.764     \\ \hline
% Shi et al. el.~\cite{corr/ShiBB17} & $768\times768$    & Titan X & 8.9  & 0.750     \\ \hline
% % TextBoxes++          & 768*768   & Titan X & 14.7      & \textbf{0.xx}     \\
% TextBoxes++          & $1024\times1024$   & Titan X & 11.6      & \textbf{0.817}     \\ \hline
% \revise{TextBoxes++\_MS}          & \revise{MS*}   & \revise{Titan X} & \revise{2.3}      & \textbf{\revise{0.829}}    \\ \hline
% \end{tabular}
% \end{table}

\begin{table}[!htbp]
\centering
\caption{Runtime and performance comparison on ICDAR 2015 Incidental Text dataset. ``F'' is short for F-measure. See corresponding text for detailed discussions.}
\label{table: speed compare}
\begin{tabular}{|c|c|c|c|c|}
\hline
Method               & Res    & FPS   & F \\ \hline
Zhang et al.~\cite{Zhang_2016_CVPR}         & MS*      & 0.476 & 0.54      \\ \hline
Tian et al.~\cite{eccv/TianHHH016}         & ss-600*      & 7.14  & 0.61      \\ \hline
Yao et al. ~\cite{corr/YaoBSZZC16}          & 480p        & 1.61  & 0.65      \\ \hline
EAST PVANET~\cite{corr/EAST}   & 720p     & \textbf{16.8}  & 0.757    \\ \hline
EAST PVANET2x~\cite{corr/EAST} & 720p     & 13.2  & 0.782     \\ \hline
EAST VGG16~\cite{corr/EAST}    & 720p     & 6.52  & 0.764     \\ \hline
Shi et al. el.~\cite{corr/ShiBB17} & $768\times768$     & 8.9  & 0.750     \\ \hline
% TextBoxes++          & 768*768   & Titan X & 14.7      & \textbf{0.xx}     \\
TextBoxes++          & $1024\times1024$    & 11.6      & \textbf{0.817}     \\ \hline
\revise{TextBoxes++\_MS}          & \revise{MS*}    & \revise{2.3}      & \textbf{\revise{0.829}}    \\ \hline
\end{tabular}
\end{table}

% \begin{figure*}[!htbp]
% \begin{center}
% \captionsetup[subfigure]{justification=centering}
%     \centering
% \subfloat[Some results on ICDAR 2013 images\label{subfig-1:spotting-visualization}]{%
%        \includegraphics[width=1.0\textwidth]{figures/end-to-end-visu1-crop.pdf}
%      }
% \hfill
% \subfloat[Some results on ICDAR 2015 Incidental text images\label{subfig-2:spotting-visualization}]{%
%        \includegraphics[width=0.99\textwidth]{figures/end-to-end-visu2-crop.pdf}
%      }
% \end{center}
% \caption{Some examples of end-to-end recognition results represented by yellow words. Note that following the evaluation protocol, Words less than 3 letters are ignored. The box colors have the same meaning as Fig.~\ref{fig:localization-visualization}.}
% \label{fig:spotting-visualization}
% \end{figure*}

\subsubsection{Runtime}
TextBoxes++ is not only accurate but also efficient. We have compared its runtime with the state-of-the-art methods on ICDAR 2015 Incidental Text dataset. As shown in Table~\ref{table: speed compare}, TextBoxes++ achieves an F-measure of 0.817 with 11.6 fps, which has a better balance on runtime and performance than the other competing methods. Note that ss-600 for the method proposed by Tian et al.~\cite{eccv/TianHHH016} means the short side of images is resized to 600. The best result on ICDAR 2015 Incidental dataset for this method is given by using a short edge of 2000, which would lead to a much slower runtime for this method. For Zhang et al.~\cite{Zhang_2016_CVPR}, MS means that they used three scales (\textit{i.e.,} 200, 500, 1000) on MSRA-TD500 dataset~\cite{Yao2012}. The method proposed in~\cite{corr/EAST} performs at 16.8 fps with PVANet~\cite{pvanet}, a faster backbone compared to VGG-16. However, the performance is 6 percents lower than TextBoxes++. To improve the performance, the authors double the number of channels of PVANet, which results in a runtime at 13.2 fps. TextBoxes++ has a similar runtime but with a 3.5 percents performance improvement. Furthermore, when the same backbone (VGG-16) is applied, the method in~\cite{corr/EAST} is much lower and still performs less well than TextBoxes++. For Shi et al.~\cite{corr/ShiBB17}, the reported runtime is tested on $768 \times 768$ MSRA-TD 500 images, but the reported performance is achieved with $720 \times 1280$ ICDAR 2015 Incidental text images. Note that the runtime for TextBoxes++ on $768 \times 768$ COCO-Text images is 19.8 fps. \revise{TextBoxes++\_MS achieves about 2.3 fps with four input scales ($384 \times 384$, $768 \times 768$, $1024 \times 1024$, $1536 \times 1536$).}

\subsection{Word spotting and end-to-end recognition to refine text detection}

\begin{table*}[!htbp]
\centering
\caption{F-measures for word spotting and end-to-end results on ICDAR 2015 Incidental Text dataset. See the corresponding dataset description in Section~\ref{subsec:datasets} for strong, weak, and generic lexicon settings. Note that the methods marked by ``*'' are published on the ICDAR 2017 Robust Reading Competition website: http://rrc.cvc.uab.es.}
 \label{ic15 end-to-end}
\begin{tabular}{|c|c|c|c|c|c|c|}
\hline
\multirow{2}{*}{Methods} & \multicolumn{3}{c|}{IC15 word spotting} & \multicolumn{3}{c|}{IC15 end-to-end} \\ \cline{2-7} 
                         & strong      & weak        & generic     & strong     & weak       & generic    \\ \hline
Megvii-Image++ *          & 0.4995      & 0.4271      & 0.3457      & 0.4674     & 0.4        & 0.3286     \\ \hline
Yunos\_Robot1.0*          & 0.4947      & 0.4947      & 0.4947      & 0.4729     & 0.4729     & 0.4729     \\ \hline
SRC-B-TextProcessingLab*  & 0.5408      & 0.5186      & 0.3712      & 0.526      & 0.5019     & 0.3579     \\ \hline
TextProposals + DictNet*  & 0.56        & 0.5226      & 0.4973      & 0.533      & 0.4961     & 0.4718     \\ \hline
Baidu IDL*                & 0.6578      & 0.6273      & 0.5165      & 0.64       & 0.6138     & 0.5071     \\ \hline
HUST\_MCLAB*              & 0.7057      & --           & --         & 0.6786     &   --         & --           \\ \hline
TextBoxes++              & \textbf{0.7645}            & \textbf{0.6904}            &  \textbf{ 0.5437}     &  \textbf{0.7334}      & \textbf{0.6587}           & \textbf{0.5190}           \\ \hline
\end{tabular}
\end{table*}

\begin{table*}[!tbp]
\centering
\caption{F-measures for word spotting and end-to-end results on ICDAR 2013 dataset. The lexicon settings are the same as for ICDAR 2015 Incidental Text dataset. %\textcolor{red}{NOte that the methods marked by ``*'' are published on the ICDAR 2015 Robust Reading Competition website: http://rrc.cvc.uab.es.}
}
\label{table:word spotting}
\renewcommand{\arraystretch}{1.2}
\begin{tabular}{|c|c|c|c|c|c|c|c|c|}
\hline
\multirow{2}{*}{Methods}    & \multirow{2}{*}{\begin{tabular}[c]{@{}c@{}}SVT\\ spotting\end{tabular}} 
& \multirow{2}{*}{\begin{tabular}[c]{@{}c@{}}SVT-50\\ spotting\end{tabular}} 
& \multicolumn{3}{c|}{\begin{tabular}[c]{@{}c@{}}IC13 \\ spotting\end{tabular}}
& \multicolumn{3}{c|}{\begin{tabular}[c]{@{}c@{}}IC13 \\ end-to-end\end{tabular}} \\ \cline{4-9} 
  &                                                                              &                                                                                 & strong                                                            & weak        & generic       & strong                      & weak                        & generic                       \\ \hline
Alsharif~\cite{alsharif2013end}        & --                                                                           & 0.48                                                                            & --                                                                  & --            & --            & --                            & --                            & --                            \\ \hline
Jaderberg~\cite{jaderberg2016reading}     & 0.56                                                                         & 0.68                                                                            & --                                                                  & --            & 0.76          & --                            & --                            & --                            \\ \hline
FCRNall+filts~\cite{gupta2016synthetic}     & 0.53                                                                         & 0.76                                                                            & --                                                                  & --            & 0.85          & --                            & --                            & --                            \\ \hline
% Deep2Text II+*   & --                                                                        & --                                                                            & 0.85                                                                 & 0.83            & 0.80          & 0.82                            & 0.79                            & 0.77                            \\ \hline
% SRC-B-TextProcessingLab*    & --                                                                        & --                                                                            & 0.90                                                                 & 0.88            & 0.81          & 0.87                            & 0.85                            & 0.80                            \\ \hline
% Adelaide\_ConvLSTMs*  & --                                                                        & --                                                                            & 0.91                                                                 & 0.90            & 0.83          & 0.87                            & 0.86                            & 0.80                            \\ \hline
TextBoxes  & \textbf{0.64} & \textbf{0.84}  & 0.94 & 0.92 & \textbf{0.87} & 0.91  & 0.89 & 0.84   \\ \hline
TextBoxes++  & \textbf{0.64} & \textbf{0.84} & \textbf{0.96} & \textbf{0.95} & \textbf{0.87} & \textbf{0.93}  & \textbf{0.92} & \textbf{0.85}   \\ \hline
\end{tabular}
\end{table*}

\begin{table}[!htbp]
\centering
\caption{Refined detection results with recognition. The evaluation method for ICDAR 2013 dataset is the IC13 Eval.
``Det'': TextBoxes++\_MS;
``Rec'': recognition without lexicon; ``Rec-lex'' : recognition with the given strong lexicon in each dataset.}
\label{table:detection+recognition}
\begin{tabular}{|c|c|c|c|c|c|c|}
\hline
\multirow{2}{*}{Datasets} & \multicolumn{3}{c|}{IC13} & \multicolumn{3}{c|}{IC15} \\ \cline{2-7} 
                          & R  & P & F & R & P & F \\ \hline
Det            & 0.844   & 0.912  & 0.876  & 0.785   & 0.878  & 0.829  \\ \hline
Det+Rec        & \textbf{0.847}   & 0.918  & 0.881  & \textbf{0.804}   & 0.881  & 0.842  \\ \hline
Det+Rec-lex    & 0.838   & \textbf{0.957}  & \textbf{0.894}  & 0.792   & \textbf{0.912}  & \textbf{0.848}  \\ \hline
\end{tabular}
\end{table}

\subsubsection{Word spotting and end-to-end recognition}
At the beginning of Section~\ref{sec: experimental results}, we have discussed the limitations of the standard text detection evaluation protocols which rely on the classical IOU threshold setting. It is meaningless to only detect text without correct recognition. In this sense, an evaluation based on the ultimate purpose of text detection would further assess the quality of text detection. For that, we have also evaluated the proposed text detector TextBoxes++ combined with a recent text recognizer CRNN model~\cite{shi2015end} in the framework of word spotting and end-to-end recognition. Note that although word spotting is similar to end-to-end recognition, the evaluation of word spotting and end-to-end recognition is slightly different. For word spotting, only some specified words are required to be detected and recognized, which implies that word spotting is generally easier than end-to-end recognition. We have tested the pipeline of TextBoxes++ followed by CRNN model~\cite{shi2015end} on three popular word spotting or end-to-end recognition benchmark datasets: ICDAR 2015 Incidental Text dataset, SVT dataset, and ICDAR 2013 dataset.

\textbf{Oriented text datasets:}
As TextBoxes++ can detect arbitrary-oriented text in natural images, we first evaluate it for word spotting and end-to-end recognition on ICDAR 2015 Incidental Text dataset. Some qualitative results are given in Fig.~\ref{fig:spotting-visualization}(a). In general, the proposed pipeline correctly recognize most oriented text. A quantitative comparison with other competing methods is depicted in Table.~\ref{ic15 end-to-end}. Note that there are not yet published papers for the competing methods in this table. These results are public on the ICDAR 2017 competition website\footnote{\label{website}\url{http://rrc.cvc.uab.es}}, and only some of the best results are shown. Our method significantly outperforms the other methods under all strong, weak, and generic lexicon for both word spotting and end-to-end recognition. More specifically, for strong lexicon, the proposed method outperforms the best competing method by 6 percent for both tasks. For weak lexicon, the proposed method improves the state-of-the-art results by 6 percents for word spotting and 4 percents for end-to-end recognition. The improvement is less significant (2.7 percents and 1.2 percents for the two tasks, respectively) when a generic lexicon is used.

%% Our method outdistance other methods with strong and weak lexicons and also lead the performance with a generic lexicon. Specifically,  our method exceeds HUST\_MCLAB* over 5 percents with in the task of strong lexicon and outperforms Baidu IDL* over 6 percents and 4 percents with weak lexicon on word spotting and end-to-end recognition respectively. The gap with generic lexicon is smaller while TextBoxes++ still runs ahead of 2.7 percents and 1.2 percents on the two tasks respectively.

\textbf{Horizontal text datasets:} We have also evaluated the proposed method for word spotting and end-to-end recognition on two horizontal text datasets: ICDAR 2013 dataset and SVT dataset. Some qualitative results on ICDAR 2013 dataset are depicted in Fig.~\ref{fig:spotting-visualization}. In general, TextBoxes++ achieves good results in various occasions, regardless of the sizes, aspect ratios, fonts, and complex backgrounds. Some quantitative results are given in Table.~\ref{table:word spotting}. 
% Note that there are not yet published papers for the methods marked by ``*'' in this table. These results are public on the ICDAR 2017 Robust Reading Competition website\footnotemark[2]. 
Our method outperforms the state-of-the-art methods. More specifically, on ICDAR 2013 dataset, our method outperforms the best competing method by at least 2 percents for all the evaluation protocols listed in Table.~\ref{table:word spotting}. The performance improvement is even more significant on SVT dataset. TextBoxes++ outperforms the state-of-the-art method~\cite{gupta2016synthetic} by at least 8 percents on both SVT and SVT-50. This is mainly because that TextBoxes++ is more robust when dealing with low-resolution images in SVT thanks to its training on relatively low-resolution images. Note that Jaderberg~\cite{jaderberg2016reading} and FCRNall+filts~\cite{gupta2016synthetic} adopt a much smaller lexicon (50k words) than our method (90k words), yet the proposed method still performs better. Compared with TextBoxes~\cite{LiaoSBWL17}, TextBoxes++ achieves better performance on ICDAR 2013 dataset.

%% Compared with the conference version TextBoxes~\cite{LiaoSBWL17}, TextBoxes++ perform better results on ICDAR 2013 dataset, owing to the multi-scale training and network refinement. However, the performance gain is ignorable on SVT because the diversity of the sizes in SVT is not as significant as ICDAR 2013 dataset.

%% More specifically, TextBoxes++ generates about 35 proposals per image when using multi-scale inputs on ICDAR 2013, with a recall of 0.93. With a strong lexicon for the recognition model, 3.8 bounding boxes per image are reserved, achieving a recall of 0.91 and a precision of 0.97.

\subsubsection{Refining detection with recognition}

We propose to use recognition to further refine detection results by integrating recognition score into detection score with Eq.~\eqref{eq:synthetic-score}. We have evaluated this idea on two datasets. As depicted in Tab.~\ref{table:detection+recognition}, the recognition without lexicon improves detection results of TextBoxes++ by 0.5 percent and 1.3 percents on ICDAR 2013 and ICDAR 2015 Incidental Text dataset, respectively. This improvement is further boosted using a specified lexicon, achieving 0.8 percent and 1.9 percents on ICDAR 2013 and ICDAR 2015 Incidental Text dataset, respectively. Note that the current text recognizer still has difficulties in dealing with vertical text and recognizing low-resolution text. A further performance improvement is expected with a better text recognizer.

% Comparing the detection results with the word spotting results on ICDAR 2013 dataset and ICDAR 2015 Incidental Text, we can see 7 percents improvement with strong lexicon on ICDAR 2013 and about 6 percents decrease with strong lexicon on ICDAR 2015 Incidental Text. The huge difference between the two datasets is caused by the recognition difficulty on the two datasets. In ICDAR 2013 dataset,  most of the text is distinct while most of the text in ICDAR 2015 Incidental Text is blurry. Besides, the text recognizer is not able to recognize vertical text. Despite of the poor recognition performance of ICDAR 2015 Incidental Text, we verified that recognition can help to refine the detection results with its semantic information. We are expecting a stronger text recognizer to further improve the results on ICDAR 2015 Incidental Text.

\subsection{Weaknesses}
\begin{figure}[!htbp]
\begin{center}
\includegraphics[width=0.9\linewidth]{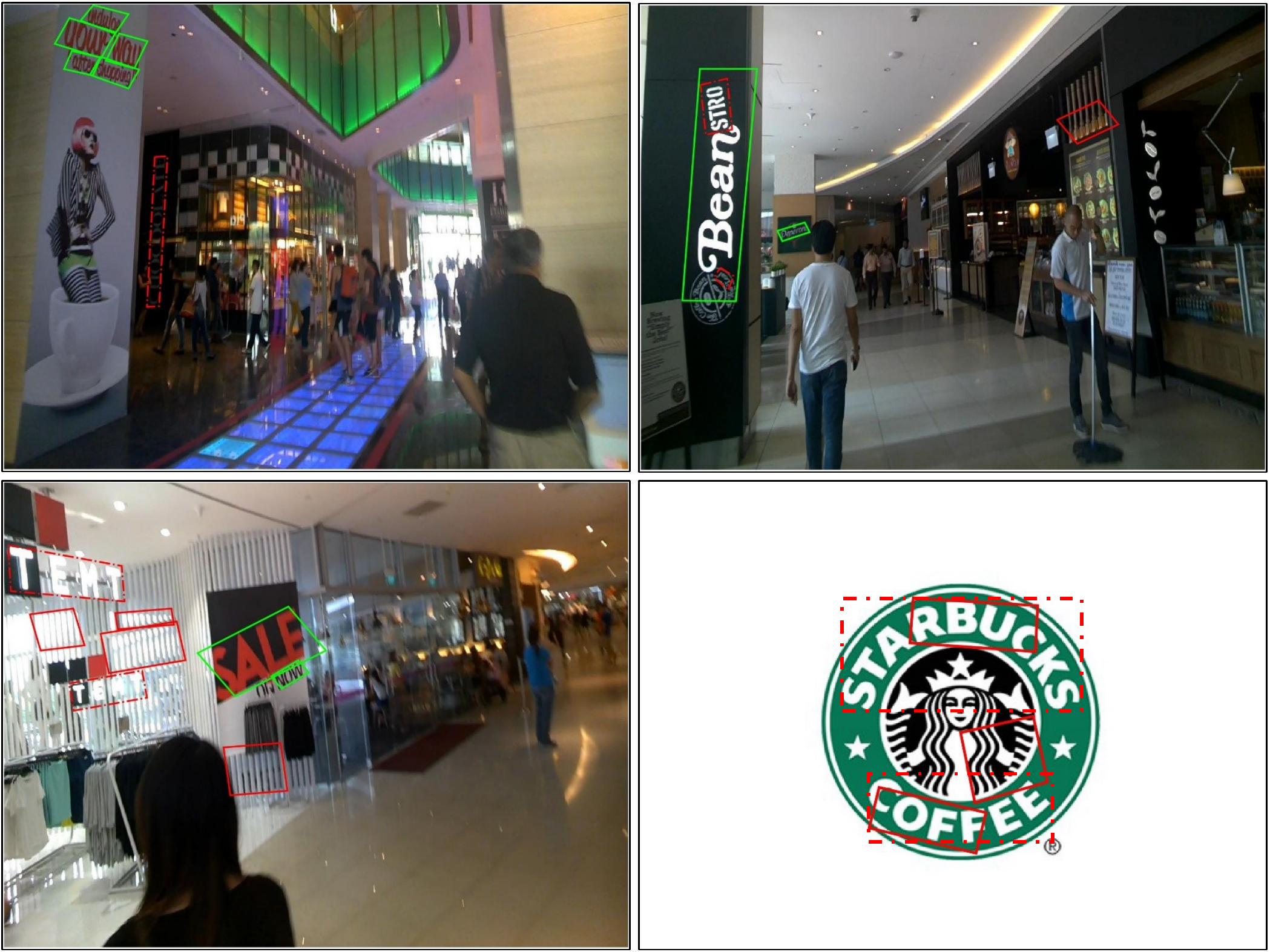}
\end{center}
% \vspace{-5mm}
\caption{\revise{Some failure examples. Green bounding boxes: correct detections; Red solid boxes: false detections; Red dashed boxes: missed ground truths.}}
\label{fig:fails}
% \vspace{-2mm}
\end{figure}
As demonstrated by previous experimental results, TextBoxes++ performs well in most situations. However, it still fails to handle some difficult cases, such as object occlusion and large character spacing. TextBoxes++ also fails to detect some vertical text due to the lack of enough vertical training data.
Even with hard negative mining, some text-like areas are still falsely detected. 
\revise{Another failure case is curved text detection. Different from some part-based methods, \textit{e.g.}~\cite{text-line-detection-based}, TextBoxes++ is hard to fit the accurate boundary for curved texts due to the limitation of quadrilateral representation.}
Note that all these difficulties also hold for the other state-of-the-art methods~\cite{corr/EAST,corr/ShiBB17}. \revise{Some failure cases are shown in Fig.~\ref{fig:fails}}.

\section{\revise{Comparisons with recent works}}
\label{sec:detailedcomparison}
\revise{We compare in detail the proposed TextBoxes++ with EAST~\cite{corr/EAST}, one of the previous state-of-the-art method, and one of the most related method DMPNet~\cite{LiuJ17b} from two aspects: simplicity and performance.}

% \revise{\textbf{TextBoxes vs. EAST}:} \revise{EAST first generates a text region map, or named as score map, using a U-shape network~\cite{u-net}. It then regresses the oriented rectangles or quadrilaterals based on the generated score map. It is a combination of segmentation and detection. In this way, it relies on pyramid-like deconvolutional layers for accurate segmentation. This extra pyramid-like deconvolutional layers requires additional computation. 
% However, TextBoxes++ directly classifies and regresses the default boxes on the convolutional feature maps, which is much simpler, avoiding the time consuming on pyramid-like deconvolution. Moreover, as compared to EAST with the same backbone (\textit{e.g.} VGG16 network), TextBoxes++ has less parameters, thus is potentially less likely to overfit. 
% For experimental results, as shown in Table. III., TextBoxes++ outperforms [40] by 3.5 percents (PVANET2x RBOX version) and 6 percents (VGG16 RBOX version) with a single scale. Moreover, as shown in Table. VI, TextBoxes++ (with a VGG-16 backbone) has a speed of 11.6fps while the VGG16 RBOX version of [40] runs at 6.52fps.}
\subsection{\revise{TextBoxes++ vs. EAST}}
% \revise{Since EAST is the previous state-of-the-art method, we compare TextBoxes++ and EAST in terms of simplicity and performance.}

\subsubsection{\revise{Simplicity}} 
\revise{
EAST generates a text region map, or named as score map, using a U-shape network~\cite{u-net}. It also regresses the oriented rectangles or quadrilaterals based on the same feature which generates the score map. 
It is a combination of segmentation and detection. In this way, it relies on pyramid-like deconvolutional layers for accurate segmentation. These extra pyramid-like deconvolutional layers require additional computation. 
However, TextBoxes++ directly classifies and regresses the default boxes on the convolutional feature maps, which is much simpler, avoiding the time consuming on pyramid-like deconvolution. This is evidenced by the speed comparison shown in Tab~\ref{table: speed compare}, where TextBoxes++ (with a VGG-16 backbone) has a speed of 11.6fps while the VGG16 RBOX version of EAST runs at 6.52fps.}

\subsubsection{\revise{Performance}} 
\revise{
EAST relies on an accurate segmentation score map as the score of the bounding boxes. Yet, the text region segmentation is challenging in itself. If the score map is not accurate enough, it is difficult to achieve correct results. For example, it is possible that the partition between two close words is predicted as text region in the segmentation score map, in this case, it is rather difficult to separate these two words in the detection. To alleviate this problem, EAST shrinks the text region in the ground truth of segmentation score map. TextBoxes++ does not suffer from such limitations. It relies on default boxes, and regresses the bounding boxes directly from the convolutional feature maps, where richer information is reserved as compared to the segmentation score map. Thus, TextBoxes++ achieves higher performance (see Table.~\ref{icdar15 text detection} ). Specifically, TextBoxes++ outperforms~\cite{LiuJ17b} by 3.5 percents (PVANET2x RBOX version) and 6 percents (VGG16 RBOX version) with a single scale.}

\subsection{\revise{TextBoxes++ vs. DMPNet}}
% \revise{We argue about the superiorities of TextBoxes++ over DMPNet in terms of simplicity and performance.}

\subsubsection{\revise{Simplicity}}
\revise{1) TextBoxes++ uses horizontal rectangles as default boxes instead of quadrilaterals with different orientations used in~\cite{LiuJ17b}. In this way, we use much fewer default boxes in every region. Furthermore, we argue that the receptive fields of the convolutional feature map are all in terms of horizontal rectangles, so the target quadrilaterals can be well regressed if it matches the receptive field. The use of oriented rectangle default boxes adopted in~\cite{LiuJ17b} is not necessary for general scene text detection. 2) Benefiting from the horizontal rectangle default boxes, TextBoxes++ enjoys a much simpler strategy for matching default boxes and ground truth by using the maximum horizontal rectangles instead of quadrilaterals. In fact, computing the intersection area between two horizontal rectangles is much easier (just using subtract operation and multiply operation once) than computing the intersection area between two arbitrary quadrilaterals, even though a Monte-Carlo method is used in~\cite{LiuJ17b}. }

\subsubsection{\revise{Performance}}
\revise{1) TextBoxes++ simultaneously regresses the maximum horizontal rectangles of the bounding boxes and the quadrilateral bounding boxes, which makes the training more stable than the method in~\cite{LiuJ17b}. 2) As compared to the method in~\cite{LiuJ17b}, TextBoxes++ goes further study on small texts in the images. We adopt a new scheme for data augmentation which is beneficial to small texts. 3) In this paper, we not only focus on scene text detection, but also concern the combination between detection and recognition. We proposed a novel score which effectively and efficiently combines the detection scores and the recognition scores.}

\revise{DMPNet~\cite{LiuJ17b} did not report the runtime. However, we argue that it is slower than TextBoxes++ based on above analysis. Moreover, TextBoxes++ outperforms DMPNet~\cite{LiuJ17b} by 11 percents (with a single scale setting) in terms of F-measure on ICDAR 2015 dataset (see Tab.~\ref{icdar15 text detection}).}

\section{Conclusion} \label{sec: conclusion}
We have presented TextBoxes++, an end-to-end fully convolutional network for arbitrary-oriented text detection, which is highly stable and efficient to generate word proposals against cluttered backgrounds. The proposed method directly predicts arbitrary-oriented word bounding boxes via a novel regression model by quadrilateral representation. The comprehensive evaluations and comparisons on some popular benchmark datasets for text detection, word spotting, and end-to-end scene text recognition, clearly validate the advantages of TextBoxes++. In all experiments, TextBoxes++ has achieved state-of-the-art performance with high efficiency for both horizontal text datasets and oriented text datasets. In the future, we plan to investigate the common failure cases (\textit{e.g.}, large character spacing and vertical text) faced by almost all state-of-the-art text detectors.

% \section{Acknowledgements}
% This work was partly supported by National Natural Science Foundation of China (61222308, 61573160, 61572207 and 61503145), and Open Project Program of the State Key Laboratory of Digital Publishing Technology (F2016001).

\bibliographystyle{IEEEtran}

\bibliography{reference}

\begin{thebibliography}{10}
\providecommand{\url}[1]{#1}
\csname url@samestyle\endcsname
\providecommand{\newblock}{\relax}
\providecommand{\bibinfo}[2]{#2}
\providecommand{\BIBentrySTDinterwordspacing}{\spaceskip=0pt\relax}
\providecommand{\BIBentryALTinterwordstretchfactor}{4}
\providecommand{\BIBentryALTinterwordspacing}{\spaceskip=\fontdimen2\font plus
\BIBentryALTinterwordstretchfactor\fontdimen3\font minus
  \fontdimen4\font\relax}
\providecommand{\BIBforeignlanguage}[2]{{%
\expandafter\ifx\csname l@#1\endcsname\relax
\typeout{** WARNING: IEEEtran.bst: No hyphenation pattern has been}%
\typeout{** loaded for the language `#1'. Using the pattern for}%
\typeout{** the default language instead.}%
\else
\language=\csname l@#1\endcsname
\fi
#2}}
\providecommand{\BIBdecl}{\relax}
\BIBdecl

\bibitem{yi2014scene}
C.~Yi and Y.~Tian, ``Scene text recognition in mobile applications by character
  descriptor and structure configuration,'' \emph{{IEEE} Trans. Image
  Processing}, vol.~23, no.~7, pp. 2972--2982, 2014.

\bibitem{xiong2016text}
B.~Xiong and K.~Grauman, ``Text detection in stores using a repetition prior,''
  in \emph{Proc. WACV}, 2016, pp. 1--9.

\bibitem{aaai/KangKY17}
C.~Kang, G.~Kim, and S.~I. Yoo, ``Detection and recognition of text embedded in
  online images via neural context models,'' in \emph{Proc. AAAI}, 2017, pp.
  4103--4110.

\bibitem{RongYT16}
X.~Rong, C.~Yi, and Y.~Tian, ``Recognizing text-based traffic guide panels with
  cascaded localization network,'' in \emph{Proc. ECCV}, 2016, pp. 109--121.

\bibitem{ye2015text}
Q.~Ye and D.~Doermann, ``Text detection and recognition in imagery: A survey,''
  \emph{IEEE TPAMI}, vol.~37, no.~7, pp. 1480--1500, 2015.

\bibitem{Pan2011}
Y.-F. Pan, X.~Hou, and C.-L. Liu, ``A hybrid approach to detect and localize
  texts in natural scene images,'' \emph{IEEE T. Image Proc.}, vol.~20, no.~3,
  pp. 800--813, 2011.

\bibitem{neumann2012real}
L.~Neumann and J.~Matas, ``Real-time scene text localization and recognition,''
  in \emph{Proc. CVPR}, 2012, pp. 3538--3545.

\bibitem{jaderberg2016reading}
M.~Jaderberg, K.~Simonyan, A.~Vedaldi, and A.~Zisserman, ``Reading text in the
  wild with convolutional neural networks,'' \emph{IJCV}, vol. 116, no.~1, pp.
  1--20, 2016.

\bibitem{bai2013scene}
B.~Bai, F.~Yin, and C.~L. Liu, ``Scene text localization using gradient local
  correlation,'' in \emph{Proc. ICDAR}, 2013, pp. 1380--1384.

\bibitem{liu2015ssd}
W.~Liu, D.~Anguelov, D.~Erhan, C.~Szegedy, and S.~E. Reed, ``{SSD:} single shot
  multibox detector,'' in \emph{Proc. ECCV}, 2016.

\bibitem{ren2015faster}
S.~Ren, K.~He, R.~Girshick, and J.~Sun, ``Faster r-cnn: Towards real-time
  object detection with region proposal networks,'' in \emph{Proc. NIPS}, 2015.

\bibitem{shi2015end}
B.~Shi, X.~Bai, and C.~Yao, ``An end-to-end trainable neural network for
  image-based sequence recognition and its application to scene text
  recognition,'' \emph{IEEE TPAMI}, vol.~39, no.~11, pp. 2298--2304, 2017.

\bibitem{LiaoSBWL17}
M.~Liao, B.~Shi, X.~Bai, X.~Wang, and W.~Liu, ``Textboxes: {A} fast text
  detector with a single deep neural network,'' in \emph{Proc. AAAI}, 2017, pp.
  4161--4167.

\bibitem{rcnn}
R.~B. Girshick, J.~Donahue, T.~Darrell, and J.~Malik, ``Rich feature
  hierarchies for accurate object detection and semantic segmentation,'' in
  \emph{Proc. CVPR}, 2014.

\bibitem{fast_rcnn}
R.~B. Girshick, ``Fast {R-CNN},'' in \emph{Proc. ICCV}, 2015.

\bibitem{RedmonDGF15}
J.~Redmon, S.~K. Divvala, R.~B. Girshick, and A.~Farhadi, ``You only look once:
  Unified, real-time object detection,'' in \emph{Proc. CVPR}, 2016.

\bibitem{selective-search}
J.~R.~R. Uijlings, K.~E.~A. van~de Sande, T.~Gevers, and A.~W.~M. Smeulders,
  ``Selective search for object recognition,'' \emph{IJCV}, vol. 104, no.~2,
  pp. 154--171, 2013.

\bibitem{svm}
M.~A. Hearst, S.~T. Dumais, E.~Osuna, J.~Platt, and B.~Scholkopf, ``Support
  vector machines,'' \emph{IEEE Intelligent Systems and their applications},
  vol.~13, no.~4, pp. 18--28, 1998.

\bibitem{Yao2012}
C.~Yao, X.~Bai, W.~Liu, Y.~Ma, and Z.~Tu, ``Detecting texts of arbitrary
  orientations in natural images,'' in \emph{Proc. CVPR}, 2012, pp. 1083--1090.

\bibitem{li2014characterness}
Y.~Li, W.~Jia, C.~Shen, and A.~van~den Hengel, ``Characterness: An indicator of
  text in the wild,'' \emph{{IEEE} Trans. Image Processing}, vol.~23, no.~4,
  pp. 1666--1677, 2014.

\bibitem{huang2014robust}
W.~Huang, Y.~Qiao, and X.~Tang, ``Robust scene text detection with convolution
  neural network induced mser trees,'' in \emph{Proc. ECCV}, 2014.

\bibitem{gomez2013multi}
L.~Gomez and D.~Karatzas, ``Multi-script text extraction from natural scenes,''
  in \emph{Proc. ICDAR}, 2013, pp. 467--471.

\bibitem{text-line-detection-based}
Y.~Guo, Y.~Sun, P.~Bauer, J.~P. Allebach, and C.~A. Bouman, ``Text line
  detection based on cost optimized local text line direction estimation,'' in
  \emph{Proc. SPIE 9395, Color Imaging XX: Displaying, Processing, Hardcopy,
  and Applications, 939507}, 2015.

\bibitem{ZhaoLK10}
M.~Zhao, S.~Li, and J.~T. Kwok, ``Text detection in images using sparse
  representation with discriminative dictionaries,'' \emph{Image Vision
  Comput.}, vol.~28, no.~12, pp. 1590--1599, 2010.

\bibitem{Zhong2016}
Z.~Zhong, L.~Jin, S.~Zhang, and Z.~Feng, ``Deeptext: A unified framework for
  text proposal generation and text detection in natural images,'' \emph{CoRR},
  vol. abs/1605.07314, 2016.

\bibitem{TextProposal}
L.~Gomez-Bigorda and D.~Karatzas, ``Textproposals: a text-specific selective
  search algorithm for word spotting in the wild,'' \emph{Pattern Recognition},
  vol.~70, pp. 60--74, 2017.

\bibitem{gupta2016synthetic}
A.~Gupta, A.~Vedaldi, and A.~Zisserman, ``Synthetic data for text localisation
  in natural images,'' in \emph{Proc. CVPR}, 2016.

\bibitem{zhang2015symmetry}
Z.~Zhang, W.~Shen, C.~Yao, and X.~Bai, ``Symmetry-based text line detection in
  natural scenes,'' in \emph{Proc. CVPR}, 2015, pp. 2558--2567.

\bibitem{long2015fully}
J.~Long, E.~Shelhamer, and T.~Darrell, ``Fully convolutional networks for
  semantic segmentation,'' in \emph{Proc. CVPR}, 2015.

\bibitem{Zhang_2016_CVPR}
Z.~Zhang, C.~Zhang, W.~Shen, C.~Yao, W.~Liu, and X.~Bai, ``Multi-oriented text
  detection with fully convolutional networks,'' in \emph{Proc. CVPR}, 2016.

\bibitem{cvpr/ChenY04}
X.~Chen and A.~L. Yuille, ``Detecting and reading text in natural scenes,'' in
  \emph{Proc. CVPR}, 2004, pp. 366--373.

\bibitem{tip/YiT12}
C.~Yi and Y.~Tian, ``Localizing text in scene images by boundary clustering,
  stroke segmentation, and string fragment classification,'' \emph{{IEEE}
  Trans. Image Processing}, vol.~21, no.~9, pp. 4256--4268, 2012.

\bibitem{eccv/TianHHH016}
Z.~Tian, W.~Huang, T.~He, P.~He, and Y.~Qiao, ``Detecting text in natural image
  with connectionist text proposal network,'' in \emph{Proc. ECCV}, 2016.

\bibitem{tip/CaoRZGF15}
X.~Cao, W.~Ren, W.~Zuo, X.~Guo, and H.~Foroosh, ``Scene text deblurring using
  text-specific multiscale dictionaries,'' \emph{{IEEE} Trans. Image
  Processing}, vol.~24, no.~4, pp. 1302--1314, 2015.

\bibitem{KangLD14}
L.~Kang, Y.~Li, and D.~S. Doermann, ``Orientation robust text line detection in
  natural images,'' in \emph{Proc. CVPR}, 2014, pp. 4034--4041.

\bibitem{matas2004robust}
J.~Matas, O.~Chum, M.~Urban, and T.~Pajdla, ``Robust wide-baseline stereo from
  maximally stable extremal regions,'' \emph{Image and vision computing},
  vol.~22, no.~10, pp. 761--767, 2004.

\bibitem{tip/YaoBL14}
C.~Yao, X.~Bai, and W.~Liu, ``A unified framework for multioriented text
  detection and recognition,'' \emph{{IEEE} Trans. Image Processing}, vol.~23,
  no.~11, pp. 4737--4749, 2014.

\bibitem{corr/YaoBSZZC16}
C.~Yao, X.~Bai, N.~Sang, X.~Zhou, S.~Zhou, and Z.~Cao, ``Scene text detection
  via holistic, multi-channel prediction,'' \emph{CoRR}, vol. abs/1606.09002,
  2016.

\bibitem{corr/ShiBB17}
B.~Shi, X.~Bai, and S.~J. Belongie, ``Detecting oriented text in natural images
  by linking segments,'' in \emph{Proc. CVPR}, 2017, pp. 3482--3490.

\bibitem{corr/EAST}
X.~Zhou, C.~Yao, H.~Wen, Y.~Wang, S.~Zhou, W.~He, and J.~Liang, ``{EAST:} an
  efficient and accurate scene text detector,'' in \emph{Proc. CVPR}, 2017, pp.
  2642--2651.

\bibitem{pvanet}
K.~Kim, Y.~Cheon, S.~Hong, B.~Roh, and M.~Park, ``{PVANET:} deep but
  lightweight neural networks for real-time object detection,'' \emph{CoRR},
  vol. abs/1608.08021, 2016.

\bibitem{LiuJ17b}
Y.~Liu and L.~Jin, ``Deep matching prior network: Toward tighter multi-oriented
  text detection,'' in \emph{Proc. CVPR}, 2017.

\bibitem{blstm}
A.~Graves and J.~Schmidhuber, ``Framewise phoneme classification with
  bidirectional {LSTM} and other neural network architectures,'' \emph{Neural
  Networks}, vol.~18, no. 5-6, pp. 602--610, 2005.

\bibitem{HeZWT17}
Z.~He, Y.~Zhou, Y.~Wang, and Z.~Tang, ``Sren: Shape regression network for
  comic storyboard extraction,'' in \emph{Pro. AAAI}, 2017, pp. 4937--4938.

\bibitem{r2cnn}
Y.~Jiang, X.~Zhu, X.~Wang, S.~Yang, W.~Li, H.~Wang, P.~Fu, and Z.~Luo,
  ``{R2CNN:} rotational region {CNN} for orientation robust scene text
  detection,'' \emph{CoRR}, vol. abs/1706.09579, 2017.

\bibitem{simonyan2014very}
K.~Simonyan and A.~Zisserman, ``Very deep convolutional networks for
  large-scale image recognition,'' \emph{CoRR}, vol. abs/1409.1556, 2014.

\bibitem{szegedy2015going}
C.~Szegedy, W.~Liu, Y.~Jia, P.~Sermanet, S.~Reed, D.~Anguelov, D.~Erhan,
  V.~Vanhoucke, and A.~Rabinovich, ``Going deeper with convolutions,'' in
  \emph{Proc. CVPR}, 2015.

\bibitem{graves2006connectionist}
A.~Graves, S.~Fern{\'a}ndez, F.~Gomez, and J.~Schmidhuber, ``Connectionist
  temporal classification: labelling unsegmented sequence data with recurrent
  neural networks,'' in \emph{Proc. ICML}, 2006, pp. 369--376.

\bibitem{icdar/KaratzasGNGBIMN15}
D.~Karatzas, L.~Gomez{-}Bigorda, A.~Nicolaou, S.~K. Ghosh, A.~D. Bagdanov,
  M.~Iwamura, J.~Matas, L.~Neumann, V.~R. Chandrasekhar, S.~Lu, F.~Shafait,
  S.~Uchida, and E.~Valveny, ``{ICDAR} 2015 competition on robust reading,'' in
  \emph{Proc. ICDAR}, 2015, pp. 1156--1160.

\bibitem{coco-text/VeitMNMB16}
A.~Veit, T.~Matera, L.~Neumann, J.~Matas, and S.~J. Belongie, ``Coco-text:
  Dataset and benchmark for text detection and recognition in natural images,''
  \emph{CoRR}, vol. abs/1601.07140, 2016.

\bibitem{karatzas2013icdar}
D.~Karatzas, F.~Shafait, S.~Uchida, M.~Iwamura, L.~G. i~Bigorda, S.~R. Mestre,
  J.~Mas, D.~F. Mota, J.~A. Almazan, and L.~P. de~las Heras, ``Icdar 2013
  robust reading competition,'' in \emph{ICDAR}, 2013, pp. 1484--1493.

\bibitem{wang2010word}
K.~Wang and S.~Belongie, ``Word spotting in the wild,'' in \emph{Proc. ECCV},
  2010, pp. 591--604.

\bibitem{adam}
D.~P. Kingma and J.~Ba, ``Adam: {A} method for stochastic optimization,''
  \emph{CoRR}, vol. abs/1412.6980, 2014.

\bibitem{AJOU}
H.~I. Koo and D.~H. Kim, ``Scene text detection via connected component
  clustering and nontext filtering,'' \emph{{IEEE} Trans. Image Processing},
  vol.~22, no.~6, pp. 2296--2305, 2013.

\bibitem{busta2015fastext}
M.~Busta, L.~Neumann, and J.~Matas, ``Fastext: Efficient unconstrained scene
  text detector,'' in \emph{Proc. ICCV}, 2015, pp. 1206--1214.

\bibitem{zamberletti2014text}
A.~Zamberletti, L.~Noce, and I.~Gallo, ``Text localization based on fast
  feature pyramids and multi-resolution maximally stable extremal regions,'' in
  \emph{Proc. ACCV}, 2014, pp. 91--105.

\bibitem{lu2015scene}
S.~Lu, T.~Chen, S.~Tian, J.-H. Lim, and C.-L. Tan, ``Scene text extraction
  based on edges and support vector regression,'' \emph{IJDAR}, vol.~18, no.~2,
  pp. 125--135, 2015.

\bibitem{tian2015text}
S.~Tian, Y.~Pan, C.~Huang, S.~Lu, K.~Yu, and C.~Lim~Tan, ``Text flow: A unified
  text detection system in natural scene images,'' in \emph{Proc. ICCV}, 2015.

\bibitem{he2016aggregating}
D.~He, X.~Yang, W.~Huang, Z.~Zhou, D.~Kifer, and C.~L. Giles, ``Aggregating
  local context for accurate scene text detection,'' in \emph{Proc. ACCV},
  2016, pp. 280--296.

\bibitem{he2016text}
T.~He, W.~Huang, Y.~Qiao, and J.~Yao, ``Text-attentional convolutional neural
  network for scene text detection,'' \emph{{IEEE} Trans. Image Processing},
  vol.~25, no.~6, pp. 2529--2541, 2016.

\bibitem{tian2017natural}
C.~Tian, Y.~Xia, X.~Zhang, and X.~Gao, ``Natural scene text detection with
  mc--mr candidate extraction and coarse-to-fine filtering,''
  \emph{Neurocomputing}, 2017.

\bibitem{qin2016fast}
S.~Qin and R.~Manduchi, ``A fast and robust text spotter,'' in \emph{Proc.
  WACV}, 2016, pp. 1--8.

\bibitem{tip/TangW17}
Y.~Tang and X.~Wu, ``Scene text detection and segmentation based on cascaded
  convolution neural networks,'' \emph{{IEEE} Trans. Image Processing},
  vol.~26, no.~3, pp. 1509--1520, 2017.

\bibitem{alsharif2013end}
O.~Alsharif and J.~Pineau, ``End-to-end text recognition with hybrid {HMM}
  maxout models,'' \emph{CoRR}, vol. abs/1310.1811, 2013.

\bibitem{u-net}
O.~Ronneberger, P.~Fischer, and T.~Brox, ``U-net: Convolutional networks for
  biomedical image segmentation,'' in \emph{Proc. MICCAI}, 2015.

\end{thebibliography}

\begin{IEEEbiography}[{\includegraphics[width=1in,height=1.25in,clip]{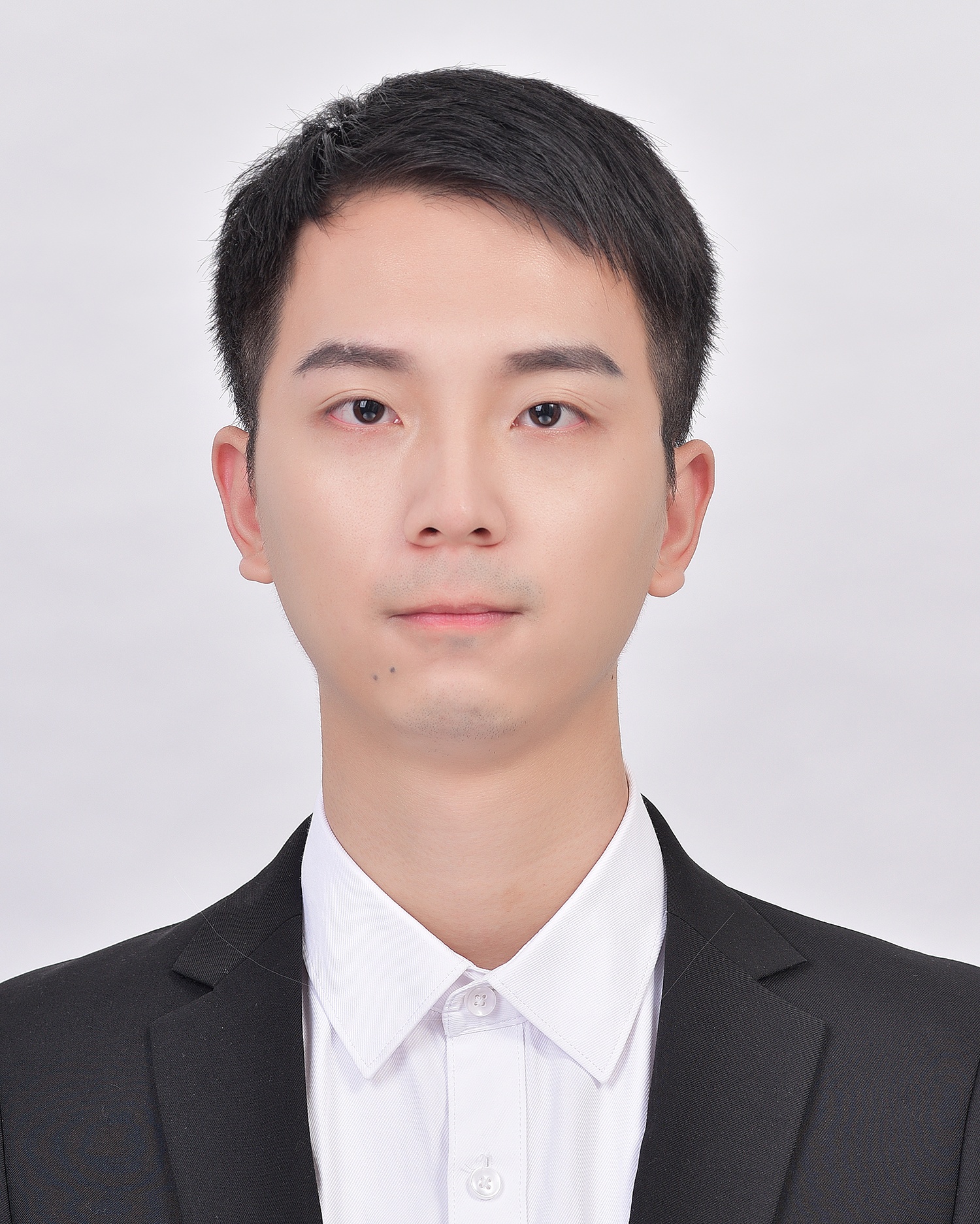}}]{Minghui Liao}
received his B.S. degree from the School of Electronic Information and Communications, Huazhong University of Science and Technology (HUST), Wuhan, China in 2016. He is currently a Ph.D. student with the School
of Electronic Information and Communications, HUST.
His research interests include scene text detection and recognition.
\end{IEEEbiography}

\begin{IEEEbiography}[{\includegraphics[width=1in,height=1.25in,clip]{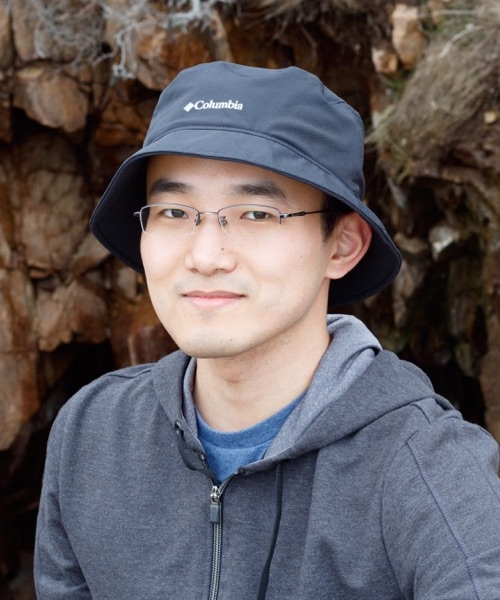}}]{Baoguang Shi}
received his B.S. degree from the School of Electronic Information and Communications, Huazhong University of Science and Technology, Wuhan, China in 2012, where he is currently a Ph.D. candidate.
He was an intern at Microsoft Research Asia in 2014, and a visiting student at Cornell University from 2016 to 2017.
His research interests include scene text detection and recognition, 3D shape recognition, and facial recognition.
\end{IEEEbiography}

\begin{IEEEbiography}[{\includegraphics[width=1in,height=1.25in,clip]{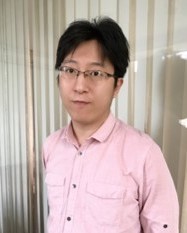}}]{Xiang Bai}
received his B.S., M.S., and Ph.D. degrees from the Huazhong University of Science and Technology (HUST), Wuhan, China, in 2003, 2005, and 2009, respectively, all in electronics and information engineering. He is currently a Professor with the School of Electronic Information and Communications, HUST. He is also the Vice-director of the National Center of Anti-Counterfeiting Technology, HUST. His research interests include object recognition, shape analysis, scene text recognition and intelligent systems. He serves as an associate editor for Pattern Recognition , Pattern Recognition Letters, Neurocomputing and Frontiers of Computer Science.
\end{IEEEbiography}

% \end{IEEEbiography}


% % insert where needed to balance the two columns on the last page with
% %\newpage



%\vfill


\end{document}